%% file: main.tex
\documentclass{article}

\usepackage{xspace}
\usepackage[dvipsnames]{xcolor}
\usepackage{pifont}

\usepackage{todonotes}
\usepackage[final]{corl_2022} %

\usepackage{booktabs}
\usepackage{colortbl}
\usepackage{graphicx}
\usepackage{enumitem}

\usepackage{wrapfig}

\usepackage{amsmath}
\usepackage{amsfonts}
\usepackage{bm}
\usepackage{vcell}

\usepackage[font=scriptsize,labelfont=sl,labelsep=period]{caption}

\newcommand{\secref}[1]{Section~\ref{#1}}
\newcommand{\appsecref}[1]{Appendix~\ref{#1}}

\newcommand{\figref}[1]{Figure~\ref{#1}}
\newcommand{\appfigref}[1]{Appendix Figure~\ref{#1}}

\newcommand{\tabref}[1]{Table~\ref{#1}}

\newcommand{\ie}{\textrm{i.e.,}\xspace}
\newcommand{\eg}{\textrm{e.g.,}\xspace}

\newcommand{\etal}{\textrm{et~al.}}

\DeclareMathOperator*{\argmax}{\arg\!\max}
\newcommand{\inputvl}{$(\mathbf{v}, \mathbf{l})$}

\definecolor{mintgreen}{RGB}{202,255,202}
\definecolor{titanwhite}{RGB}{238,238,255}

\usepackage{todonotes}

\newcommand{\bcz}{Image-BC~}
\newcommand{\bczcnn}{Image-BC (CNN)~}
\newcommand{\bczvit}{Image-BC (ViT)~}
\newcommand{\unet}{C2FARM-BC~}

\newcommand{\highlight}[1]{\textcolor{black}{#1}}

\newcommand{\model}{\textsc{PerAct}}
\newcommand{\modelfullns}{\textsc{Perceiver-Actor}}

\newcommand{\modelfull}{\textsc{\modelfullns~}}

\newcommand{\blank}{\rule{0.3cm}{0.25mm}~}

\vspace{-1cm}
\title{\modelfullns: \\ A Multi-Task Transformer for Robotic Manipulation}

\vspace{-2cm}
\author{ Mohit Shridhar~$^{1,}$\thanks{Work done partly while the author was a part-time intern at NVIDIA.}%
 \hspace{8px} Lucas Manuelli~$^2$ \hspace{4px}  Dieter Fox~$^{1, 2}$\\
$^1$University of Washington \hspace{6px} $^2$NVIDIA\\
\tt\small mshr@cs.washington.edu \hspace{2px} lmanuelli@nvidia.com \hspace{2px} fox@cs.washington.edu\\[1em]
\large\textbf{\url{peract.github.io}}
}

\begin{document}
\maketitle

\vspace{-1.0cm}
\begin{abstract}
    Transformers have revolutionized vision and natural language processing with their ability to scale with large datasets. But in robotic manipulation, data is both limited and expensive.  
    Can manipulation still benefit from Transformers with the right problem formulation?
    We investigate this question with \model, a language-conditioned behavior-cloning agent for multi-task 6-DoF manipulation. 
    \model~encodes language goals and RGB-D voxel observations with a Perceiver Transformer~\citep{jaegle2021perceiver}, and outputs discretized actions by ``detecting the next best voxel action''. 
    Unlike frameworks that operate on 2D images, the voxelized 3D  observation and action space provides a strong structural prior for efficiently learning 6-DoF actions.
    With this formulation, we train a single multi-task Transformer for \highlight{18 RLBench tasks (with 249 variations) and 7 real-world tasks (with 18 variations)} from just a few demonstrations per task.
    Our results show that \model~significantly outperforms unstructured image-to-action agents and 3D ConvNet baselines for a wide range of tabletop tasks.

\end{abstract}

\keywords{Transformers, Language Grounding, Manipulation, Behavior Cloning}

\section{Introduction}

Transformers~\citep{vaswani2017attention} have become  prevalent in natural language processing and computer vision. By framing problems as sequence modeling tasks, and training on large amounts of diverse data, Transformers have achieved groundbreaking results in several domains~\citep{brown2020language,dosovitskiy2020image,jumper2021highly,vinyals2019alphastar}. Even in domains that do not conventionally involve sequence modeling~\citep{chen2021pix2seq,chen2021decision}, Transformers have been adopted as a \textit{general} architecture~\citep{reed2022generalist}. But in robotic manipulation, data is both limited and expensive. 
Can we still bring the power of Transformers to 6-DoF manipulation with the right problem formulation?

Language models operate on sequences of tokens~\citep{devlin2018bert}, and vision transformers operate on sequences of image patches~\citep{dosovitskiy2020image}. While pixel transformers~\citep{jaegle2021perceivericml,jaegle2021perceiver} exist, they are not as data efficient as approaches that use convolutions or patches to exploit the 2D structure of images. Thus, while Transformers may be domain agnostic, they still require the right problem formulation to be data efficient. 
A similar efficiency issue is apparent in behavior-cloning (BC) agents that directly map 2D images to 6-DoF actions.
\highlight{Agents like Gato~\citep{reed2022generalist} and BC-Z~\citep{jang2022bc,ahn2022can} have shown impressive multi-task capabilities, but they require several weeks or even months of data collection}.
\highlight{In contrast, recent works in reinforcement-learning like C2FARM~\citep{c2farm} construct a voxelized  observation and action space to efficiently learn visual representations of 3D actions with 3D ConvNets.}  
Similarly, in this work, we aim to exploit the 3D structure of \textit{voxel patches} for efficient 6-DoF behavior-cloning with Transformers (analogous to how vision transformers~\citep{dosovitskiy2020image} exploit the 2D structure of image patches).

\begin{figure*}[!t]
    \centering
    \hspace*{-1.45cm}
    \includegraphics[width=1.2\textwidth]{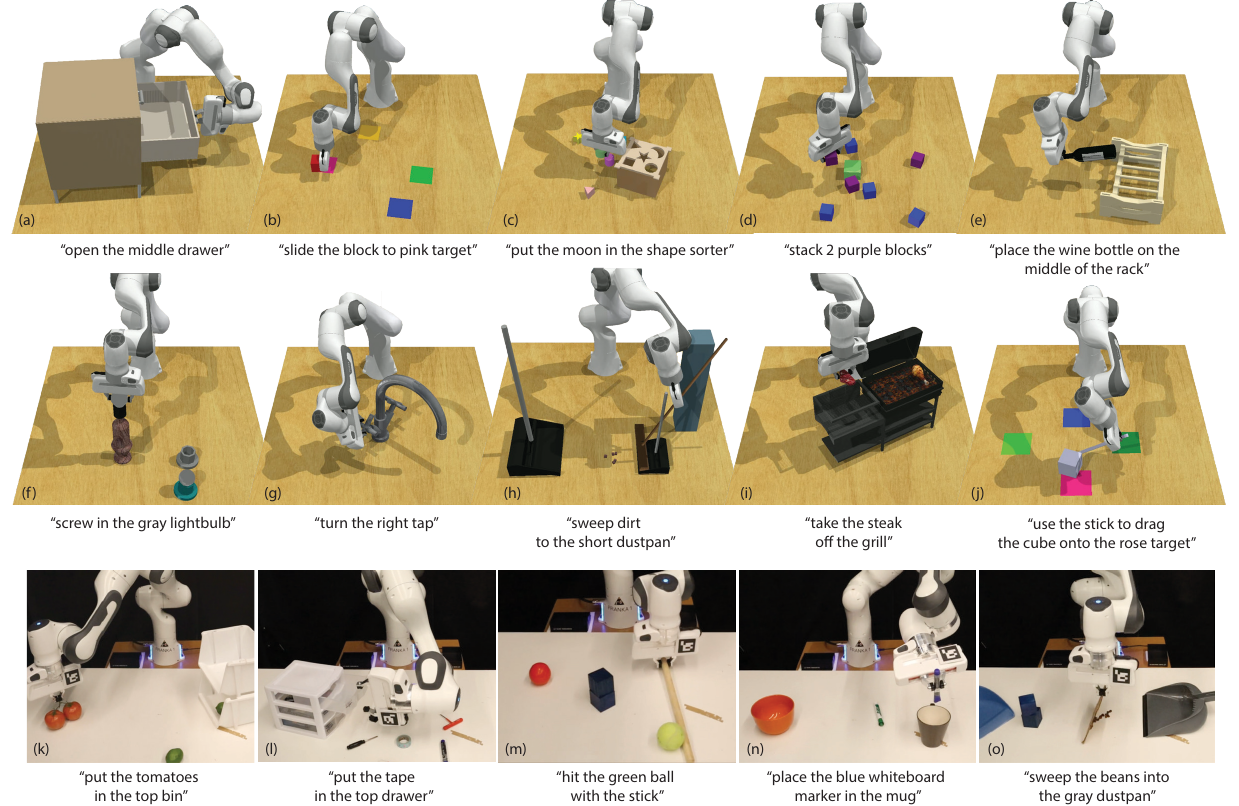}
    \caption{\textbf{Language-Conditioned Manipulation  Tasks:} \model~is a language-conditioned multi-task agent capable of imitating a wide range of 6-DoF manipulation tasks. We conduct experiments on 18 simulated tasks in RLBench~\citep{james2020rlbench} (a-j; only 10 shown), with several pose and semantic variations. We also demonstrate our approach with a Franka Panda on 7 real-world tasks (k-o; only 5 shown) with a multi-task agent trained with just 53 demonstrations. See the supplementary video for simulated and real-world rollouts.}
    \label{fig:tasks}
    \vspace{-1.5em}
\end{figure*}

To this end, we present \model~(short for \modelfullns), a language-conditioned BC agent that can learn to imitate a wide variety of 6-DoF manipulation tasks with just a few demonstrations per task. \model~encodes a sequence of RGB-D voxel patches and predicts discretized translations, rotations, and gripper actions that are executed with a motion-planner in an observe-act loop. \model~is essentially a classifier trained with supervised learning to \textit{detect actions} akin to prior work like CLIPort~\citep{cliport,zengTransporterNetworksRearranging2021},
except our observations and actions are represented with 3D voxels instead of 2D image pixels. 
Voxel grids are  less prevalent than images in end-to-end BC approaches often due to scaling issues with high-dimensional inputs. But in \model, we use a Perceiver\footnote{Throughout the paper we refer to PerceiverIO~\citep{jaegle2021perceiver} as Perceiver for brevity.} Transformer~\citep{jaegle2021perceiver} to encode very high-dimensional input of up to 1 million voxels with only a small set of latent vectors. This voxel-based formulation provides a strong structural prior with several benefits: a natural method for fusing multi-view observations, learning robust action-centric\footnote{\highlight{Action-centric refers to a  system that learns perceptual representations of actions; see \appsecref{app:more_qpred}.}} representations~\citep{gibson2014ecological,brooks1991new}, and enabling data augmentation in 6-DoF -- all of which help learn generalizable skills by focusing on \textit{diverse} rather than narrow multi-task data.  

To study the effectiveness of this formulation, we conduct large-scale experiments in the RLBench~\citep{james2020rlbench} environment. We train a single multi-task agent on 18 diverse tasks with 249 variations that involve a range of prehensile and non-prehensile behaviors like placing wine bottles on a rack and dragging objects with a stick (see \figref{fig:tasks} a-j). Each task also includes several pose and semantic variations with objects that differ in placement, color, shape, size, and category. 
Our results show that \model~significantly outperforms image-to-action agents \highlight{(by $34\times$)} and 3D ConvNet baselines \highlight{(by $2.8\times$)}, without using any explicit representations of instance segmentations, object poses, memory, or symbolic states. 
We also validate our approach on a Franka Panda with a multi-task agent trained \textit{from scratch} on 7 real-world tasks  with a \textbf{total of just 53 demonstrations} (see \figref{fig:tasks} k-o).

In summary, our contributions are as follows:
\vspace{-0.2em}
\begin{itemize}[leftmargin=0.8cm,itemsep=0.05em]
    \item A \textbf{novel problem formulation} for perceiving, acting, and specifying goals with Transformers.
    \item An efficient \textbf{action-centric} framework for \textbf{grounding language in 6-DoF actions}.
    \item \textbf{Empirical results} investigating multi-task agents on a range of simulated and real-world tasks. 
\end{itemize}
The code and pre-trained models will be made available  at \url{peract.github.io}.

\section{Related Work}
\label{sec:related_work}
\textbf{Vision for Manipulation.} Traditionally, methods in robot perception have used explicit ``object'' representations like instance segmentations, object classes, poses  \citep{he2017mask,Xiang-RSS-18,zhu2014single,zeng2017multi,deng2020self,xie2020best}. Such methods struggle with deformable and granular items like cloths and beans that are hard to represent with geometric models or segmentations. 
In contrast, recent methods~\citep{zeng2019robotic,zengTransporterNetworksRearranging2021,cliport,stengel2022guiding} learn action-centric representations without any ``objectness'' assumptions, but they are limited to top-down 2D settings with simple pick-and-place primitives. \highlight{In 3D, James et al. proposed C2FARM~\citep{c2farm}, an action-centric reinforcement learning (RL) agent with a coarse-to-fine-grain 3D-UNet backbone. The coarse-to-fine-grain scheme has a limited receptive field that cannot look at the entire scene at the finest level. In contrast, \model~learns action-centric representations with a global-receptive field through a Transformer backbone. Also, \model~does BC instead of RL, which enables us to easily train a multi-task agent for several tasks by conditioning it with language goals}. 

\textbf{End-to-End Manipulation} approaches~\citep{kalashnikov2018qt,Wu-RSS-20,levine2016end,finn2017deep} make the least assumptions about objects and tasks, but are often formulated as an image-to-action prediction task. Training directly on RGB images for 6-DoF tasks is often inefficient, generally requiring several  demonstrations or episodes just to learn basic skills like rearranging objects. In contrast, \model~uses a voxelized observation and action space, which is dramatically more efficient and robust in 6-DoF settings.
While other works in 6-DoF grasping~\citep{song2020grasping,murali20206,mousavian20196,xu2022umpnet,agrawal2021scene,simeonov2021neural} have used RGB-D and pointcloud input, they have not been applied to sequential tasks or used with language-conditioning.
Another line of work tackles data inefficiency by using pre-trained image representations ~\citep{cliport,nair2022r3m,yuan2021sornet} to bootstrap BC. Although our framework is trained from scratch, such pre-training approaches could be integrated together in future works for even greater efficiency and generalization to unseen objects. 

\textbf{Transformers for Agents and Robots.} Transformers have become the prevalent architecture in several domains. Starting with NLP~\citep{vaswani2017attention,brown2020language,liu2019roberta}, recently in vision~\citep{dosovitskiy2020image,liu2021swin}, and even RL~\citep{chen2021decision,janner2021offline,lee2022multi}. In robotics, Transformers have been applied to assistive teleop~\citep{clever2021assistive},  legged locomotion~\citep{yang2021learning},  path-planning~\citep{chaplot2021differentiable,johnson2021motion}, imitation learning~\citep{dasari2020transformers,kim2021transformer}, morphology controllers~\citep{gupta2022metamorph}, spatial rearrangement~\citep{liu2022structformer}, and grasping~\citep{han2021learning}. Transformers have also achieved impressive results in multi-domain settings like in Gato~\cite{reed2022generalist} where a single Transformer was trained on 16 domains such as captioning, language-grounding, robotic control etc. However, Gato relies on extremely large datasets like 15K episodes for block stacking and 94K episodes for Meta-World~\citep{yu2020meta} tasks. Our approach might complement agents like Gato, which could use our 3D formulation for greater efficiency and robustness. 

\textbf{Language Grounding for Manipulation.} Several works have proposed methods for grounding language in robot actions~\citep{Shridhar-RSS-18,matuszek2014learning,bollini2013interpreting,misra2016tell,bisk2016natural,thomason2015learning,interact_picking18,chenJointNetworkGrasp2021, blukis2020few, paxton2019prospection,tellex2011understanding,nguyen2020robot}. However, these methods use disentangled pipelines for perception and action, with the language primarily being used to guide perception~\citep{egl}. 
Recently, a number of end-to-end approaches \citep{ahn2022can,jang2022bc,nair2022learning,mees2022matters,lynch2020grounding} have been proposed for conditioning BC agents with language instructions. These methods require thousands of human demos or autonomous episodes that are collected over several days or even months. In contrast, \model~can learn robust multi-task policies with just a few minutes of training data. For benchmarking, several simulation environments exist~\citep{mees2021calvin,zengTransporterNetworksRearranging2021,yu2020meta}, but we use RLBench~\citep{james2020rlbench} for its diversity of 6-DoF tasks and ease of generating demonstrations with templated language goals. 

\section{\modelfull}

\begin{figure*}[!t]
    \centering
    \hspace*{-2.2cm}
    \includegraphics[width=1.3\textwidth]{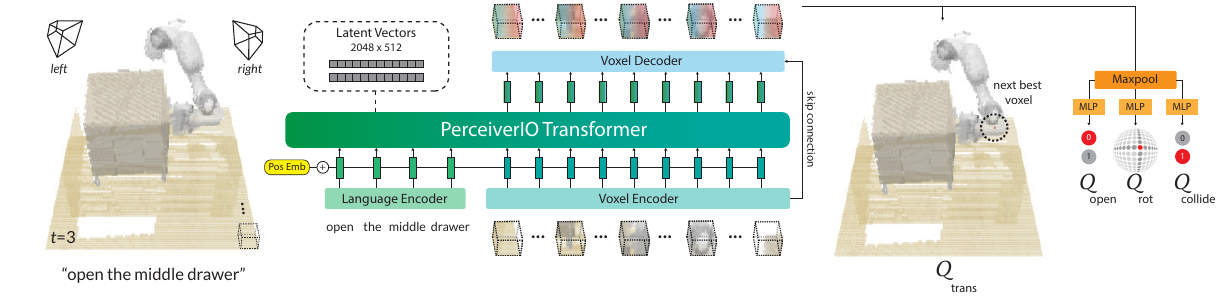}
    \vspace{-0.4cm}
    \caption{\textbf{\model~Overview.} \model~is a language-conditioned behavior-cloning agent  trained with supervised learning to \textit{detect actions}. \model~takes as input a language goal and a voxel grid reconstructed from RGB-D sensors. The voxels are split into 3D patches, and the language goal is encoded with a pre-trained language model. These language and voxel features are appended together as a sequence and encoded with a Perceiver transformer~\citep{jaegle2021perceiver}. 
    Despite the extremely long input sequence, Perceiver uses a small set of latent vectors to encode the input (see \appfigref{fig:perceiver_arch} for an illustration). 
    These encodings are upsampled back to the original voxel dimensions with a decoder and reshaped with linear layers to predict a discretized translation, rotation, gripper open, and collision avoidance action. This action is executed with a motion-planner after which the new observation is used to predict the next discrete action in an observe-act loop until termination. 
    }
    \label{fig:peract}
    \vspace{-1.5em}
\end{figure*}

\model~is a language-conditioned behavior-cloning agent for 6-DoF manipulation. The key idea is to learn perceptual representations of actions conditioned on language goals. Given a voxelized reconstruction of a scene, we use a Perceiver Transformer~\citep{jaegle2021perceiver} to learn per-voxel features. Despite the extremely large input space ($100^3$), Perceiver uses a small set of latent vectors to encode the input. The per-voxel features are then used to predict the next best action in terms of discretized translation, rotation, and gripper state at each timestep. \model~relies purely on the current observation to determine what to do next in sequential tasks. 
See \figref{fig:peract} for an overview.

\secref{sec:demos} and \secref{sec:keyframe_and_voxel} describe our dataset setup. \secref{sec:peract} describes our problem formulation with \model, and \secref{sec:training} provides details on training \model. Further implementation details are presented in \appsecref{app:peract_details}. 

\newpage
 
\vspace{-0.3cm} 
\subsection{Demonstrations} \label{sec:demos}
\vspace{-0.2cm}
We assume access to a dataset  $\mathcal{D} = \{\zeta_1, \zeta_2, \ldots, \zeta_n\}$ of $n$ expert demonstrations, each paired with English language goals $\mathcal{G} = \{\mathbf{l}_{1}, \mathbf{l}_{2}, \ldots, \mathbf{l}_{n}\}$. 
These demonstrations are collected by an expert with the aid of a motion-planner to reach intermediate poses. Each demonstration $\zeta$ is a sequence of continuous actions $\mathcal{A} = \{a_{1}, a_{2}, \ldots, a_{t}\}$ paired with observations $\mathcal{O} = \{\Tilde{o}_{1}, \Tilde{o}_{2}, \ldots \Tilde{o}_{t}\}$. An action $a$ consists of the 6-DoF pose, gripper open state, and whether the motion-planner used collision avoidance to reach an intermediate pose: $a = \{a_{\textrm{pose}}, a_{\textrm{open}}, a_{\textrm{collide}}\}$. 
An observation $\Tilde{o}$ consists of RGB-D images from any number of cameras. We use four cameras for simulated experiments $\Tilde{o}_{\textrm{sim}} = \{o_{\textrm{front}}, o_{\textrm{left}}, o_{\textrm{right}}, o_{\textrm{wrist}} \}$, but just a single camera  for real-world experiments $\Tilde{o}_{\textrm{real}} = \{o_{\textrm{front}} \}$.  
\vspace{-0.2cm}
\subsection{Keyframes and Voxelization}
\label{sec:keyframe_and_voxel}
\vspace{-0.2cm}

Following prior work by James et al.~\citep{c2farm}, we construct a  structured observation and action space through keyframe extraction and voxelization.

Training our agent to directly predict continuous actions is inefficient and noisy. So instead, for each demonstration $\zeta$, we extract a set of keyframe actions $\{\mathbf{k}_{1}, \mathbf{k}_{2}, \ldots, \mathbf{k}_{m}\} \subset \mathcal{A}$ \highlight{that capture bottleneck end-effector poses}~\citep{johns2021coarse} in the action sequence with a simple heuristic: an action is a keyframe if (1) the joint-velocities are near zero and (2) the gripper open state has not changed. 
Each datapoint in the demonstration $\zeta$ can then be cast as a \highlight{``predict the next (best) keyframe action'' task~\citep{c2farm,armpaper,liu2022auto_lambda}}. See \appfigref{app:keypoints_and_demo} for an illustration of this process.

To learn action-centric representations~\cite{gibson2014ecological} in 3D, we use a voxel grid~\citep{moravec1996robot,30724} to represent both the observation and action space. The observation voxels $\mathbf{v}$ are reconstructed from RGB-D observations $\Tilde{o}$ fused through triangulation $\Tilde{o} \Rightarrow \mathbf{v}$ from known camera extrinsics and intrinsics. By default, we use a voxel grid of $100^3$, which corresponds to a volume of $1.0\textrm{m}^3$ in metric scale. The keyframe actions $\mathbf{k}$ are discretized such that training our BC agent can be formulated as a \highlight{``next best action'' classification task~\citep{c2farm}}.
Translation is simply the closest voxel to the center of the gripper fingers. Rotation is discretized into 5 degree bins for each of the three rotation axes. Gripper open state is a binary value. Collide is also a binary value that indicates if the motion-planner should avoid everything in the voxel grid or nothing at all; switching between these two modes of collision avoidance is crucial as tasks often involve both contact based (\eg pulling the drawer open) and non-contact based motions (\eg reaching the handle without colliding into anything). 
\vspace{-0.2cm}

\subsection{\model~Agent} \label{sec:peract}
\vspace{-0.2cm}
\model~is a Transformer-based~\citep{vaswani2017attention} agent that takes in a voxel observation and language goal \inputvl, and outputs a discretized translation, rotation, and gripper open action. This action is executed with a motion-planner, after which this process is repeated until the goal is reached.

The language goal $\mathbf{l}$ is encoded with a pre-trained language model. 
We use CLIP's~\citep{radfordLearningTransferableVisual2021} language encoder, but any pre-trained language model would suffice~\citep{ahn2022can,lynch2020grounding}. Our choice of CLIP opens up possibilities for future work to use pre-trained vision features that are aligned with the language for better generalization to unseen semantic categories and instances~\citep{cliport}. 

The voxel observation $\mathbf{v}$ is split into 3D patches of size $5^3$ (akin to vision-transformers like ViT~\citep{dosovitskiy2020image}). In implementation, these  patches are extracted with a 3D convolution layer with a kernel-size and stride of 5, and then flattened into a sequence of voxel encodings. The language encodings are fine-tuned with a linear layer and then appended with the voxel encodings to form the input sequence. We also add learned positional embeddings to the sequence to incorporate voxel and token positions.

The input sequence of language and voxel encodings is extremely long. 
A standard Transformer with $\mathcal{O}(n^2)$ self-attention connections and an input of  $(100/5)^3 = 8000$ patches is hard to fit on the memory of a commodity GPU.  
Instead, we use the Perceiver~\citep{jaegle2021perceiver} Transformer. Perceiver is a latent-space Transformer, where instead of attending to the entire input, it first computes cross-attention between the input and a much smaller set of latent vectors (which are randomly initialized and trained). These latents are encoded with self-attention layers, and for the final output, the latents are again cross-attended with the input to match the input-size. See \appfigref{fig:perceiver_arch} for an illustration. By default, we use $2048$ latents of dimension 512 : $\mathbb{R}^{2048 \times 512}$, but in \appsecref{app:ablations} we experiment with different latent sizes. 

\begin{figure*}[!t]
    \centering
    \vspace{-1.2cm}
    \hspace*{-1.5cm}
    \includegraphics[width=1.2\textwidth]{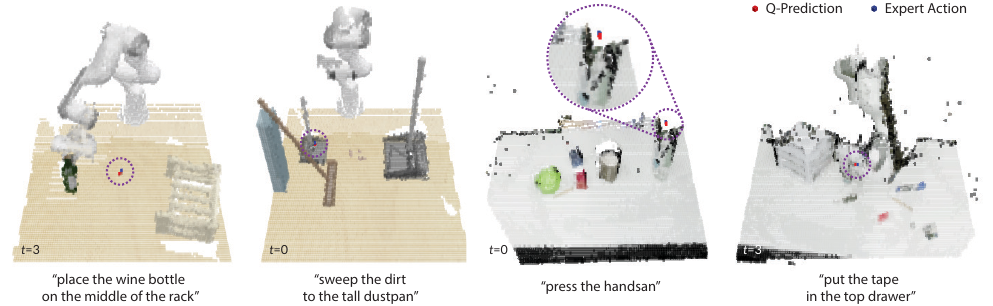}
    \caption{\textbf{Q-Prediction Examples}: Qualitative examples of translation $\mathcal{Q}$-Predictions from \model~along with expert actions, highlighted with dotted-circles. The left two are simulated tasks, and the right two are real-world tasks. See \appsecref{app:more_qpred} for more examples.}
    \label{fig:q_pred} 
    \vspace{-1em}
\end{figure*}

The Perceiver Transformer uses 6 self-attention layers to encode the latents and outputs a sequence of patch encodings from the output cross-attention layer. These patch encodings are upsampled with a 3D convolution layer and tri-linear upsampling to decode 64-dimensional voxel features. The decoder includes a skip-connection from the encoder (like in UNets~\citep{ronneberger2015u}). The per-voxel features are then used to predict discretized actions~\citep{c2farm}. For translation, the voxel features are reshaped into the original voxel grid ($100^3$) to form a 3D $\mathcal{Q}$-function of action-values. For rotation, gripper open, and collide, the features are max-pooled and then decoded with linear layers to form their respective $\mathcal{Q}$-function. The best action $\mathcal{T}$ is chosen by simply maximizing the $\mathcal{Q}$-functions:
\begin{align*}
\mathcal{T}_{\textrm{trans}} = \underset{(x,y,z)}{\argmax} \  \mathcal{Q}_{\textrm{trans}}((x,y,z) \ | \ \mathbf{v}, \mathbf{l} \,), \hspace{1cm}  &
\mathcal{T}_{\textrm{rot}} = \underset{(\psi,\theta,\phi)}{\argmax} \  \mathcal{Q}_{\textrm{rot}}((\psi,\theta,\phi) \ | \ \mathbf{v}, \mathbf{l} \,), \\
\mathcal{T}_{\textrm{open}} = \underset{\omega}{\argmax} \  \mathcal{Q}_{\textrm{open}}(\, \omega \ | \ \mathbf{v}, \mathbf{l} \,), \hspace{1cm}  &
\mathcal{T}_{\textrm{collide}} = \underset{\kappa}{\argmax} \ \mathcal{Q}_{\textrm{collide}}(\, \kappa \ | \ \mathbf{v}, \mathbf{l} \,),
\end{align*}
where $(x, y, z)$ is the voxel location in the grid, $(\psi, \theta, \phi)$ are discrete rotations in Euler angles, $\omega$ is the gripper open state and $\kappa$ is the collide variable. See \figref{fig:q_pred} for examples of  $\mathcal{Q}$-predictions.

\vspace{-0.1cm}
\subsection{Training Details}
\label{sec:training}
\vspace{-0.1cm}

\model~is trained through supervised learning with discrete-time input-action tuples from a dataset of demonstrations. These tuples are composed of voxel observations, language goals, and keyframe actions $\{(\mathbf{v}_{1}, \mathbf{l}_{1}, \mathbf{k}_{1}), (\mathbf{v}_{2}, \mathbf{l}_{2}, \mathbf{k}_{2}), \ldots\}$. During training, we randomly sample a tuple and supervise the agent to predict the keyframe action $\mathbf{k}$ given the observation and goal \inputvl. For translation, the ground-truth action is represented as a one-hot voxel encoding $Y_{\textrm{trans}} : \mathbb{R}^{H \times W \times D}$. Rotations are also represented with a one-hot encoding per rotation axis with $R$ rotation bins $Y_{\textrm{rot}} : \mathbb{R}^{(360/R) \times 3}$  ($R=5$ degrees for all experiments). Similarly, open and collide variables are binary one-hot vectors $Y_{\textrm{open}} : \mathbb{R}^{2}$, $Y_{\textrm{collide}} : \mathbb{R}^{2}$. The agent is trained with cross-entropy loss like a  classifier: 
\begin{equation*}
    \mathcal{L}_{\textrm{total}} = - \mathbb{E}_{Y_{\textrm{trans}}}[\textrm{log} \mathcal{V}_{\textrm{trans}}] - \mathbb{E}_{Y_{\textrm{rot}}}[\textrm{log} \mathcal{V}_{\textrm{rot}}] - \mathbb{E}_{Y_{\textrm{open}}}[\textrm{log} \mathcal{V}_{\textrm{open}}] -
    \mathbb{E}_{Y_{\textrm{collide}}}[\textrm{log} \mathcal{V}_{\textrm{collide}}],
\end{equation*}
where $\mathcal{V}_{\textrm{trans}} = \textrm{softmax}(\mathcal{Q}_{\textrm{trans}}((x,y,z) | \mathbf{v}, \mathbf{l}))$, $\mathcal{V}_{\textrm{rot}} = \textrm{softmax}(\mathcal{Q}_{\textrm{rot}}((\psi, \theta, \phi) | \mathbf{v}, \mathbf{l}))$, $\mathcal{V}_{\textrm{open}} = \textrm{softmax}(\mathcal{Q}_{\textrm{open}}(\omega | \mathbf{v}, \mathbf{l}))$, $\mathcal{V}_{\textrm{collide}} = \textrm{softmax}(\mathcal{Q}_{\textrm{collide}}(\kappa | \mathbf{v}, \mathbf{l}))$ respectively. For robustness, we also augment $\mathbf{v}$ and $\mathbf{k}$ with translation and rotation perturbations. See \appsecref{app:data_aug} for more details.

By default, we use a voxel grid size of $100^3$. We conducted validation tests by replaying expert demonstrations with discretized actions to ensure that $100^3$ is a sufficient resolution for execution. The agent was trained with a batch-size of 16 on 8 NVIDIA V100 GPUs for 16 days (600K iterations). We use the LAMB~\citep{you2019large} optimizer following Perceiver~\citep{jaegle2021perceiver}.

For multi-task training, we simply sample input-action tuples from all tasks in the dataset. To ensure that tasks with longer horizons are not over-represented during sampling, each batch contains a uniform distribution of tasks. That is, we first uniformly sample a set of tasks of batch-size length, then pick a random input-action tuple for each of the sampled tasks. With this strategy, longer-horizon tasks need more training steps for full coverage of input-action pairs, but all tasks are given equal weighting during gradient updates.

\section{Results}
\vspace{-0.2cm}
We perform experiments to answer the following questions: (1) How effective is \model~compared to unstructured image-to-action frameworks and standard architectures like 3D ConvNets? And what are the factors that affect \model's performance? (2) Is the global receptive field of Transformers actually beneficial over methods with local receptive fields? (3) Can \model~be trained on real-world tasks with noisy data?

\vspace{-0.2cm}
\subsection{Simulation Setup}
\vspace{-0.2cm}

We conduct our primary experiments in simulation for the sake of reproducibility and benchmarking.

\textbf{Environment.} The simulation is set in CoppelaSim~\citep{coppelasim} and interfaced through PyRep~\citep{james2019pyrep}. All experiments use a Franka Panda robot with a parallel gripper. The input observations are captured from four RGB-D cameras positioned at the front, left shoulder, right shoulder, and on the wrist, as shown in \appfigref{fig:sim_setup}. All cameras are noiseless and have a resolution of $128 \times 128$. 

\textbf{Language-Conditioned Tasks.}  \highlight{We train and evaluate on 18 RLBench~\citep{james2020rlbench} tasks. See \href{https://peract.github.io}{peract.github.io} for examples and \appsecref{app:task_details} for details on individual tasks. Each task includes several variations, ranging from 2-60 possibilities, \eg in the \texttt{stack blocks} task, \textit{``stack 2 red blocks''} and \textit{``stack 4 purple blocks''} are two variants}. These variants are randomly sampled during data generation, but kept consistent during evaluations for one-to-one comparisons. 
Some RLBench tasks were modified to include additional variations to stress-test multi-task and language-grounding capabilities. 
There are a total of 249 variations across 18 tasks, and the number of extracted keyframes range from 2-17.
\highlight{All keyframes from an episode have the same language goal, which is constructed from templates} (but human-annotated for real-world tasks). Note that in all experiments, we do not test for generalization to unseen objects, \ie~our train and test objects are the same. 
\highlight{However during test time, the agent has to handle novel object poses, randomly sampled goals, and randomly sampled scenes with different semantic instantiations of object colors, shapes, sizes, and categories}.
The focus here is to evaluate the performance of a single multi-task agent trained on all tasks and variants.

\textbf{Evaluation Metric.} Each multi-task agent is evaluated independently on all 18 tasks. Evaluations are scored either 0 for failures or 100 for complete successes. There are no partial credits. We report average success rates on 25 evaluation episodes per task ($25 \times18 = 450$ total episodes) for agents trained with $n=10,100$ demonstrations per task. During evaluation, an agent keeps taking actions until an oracle indicates task-completion or reaches a maximum of 25 steps.

\vspace{-0.2cm}
\subsection{Simulation Results}
\label{sec:sim_results}
\vspace{-0.1cm}

\tabref{table:rlbench} reports success rates of multi-task agents trained on all 18 tasks. 
We could not investigate single-task agents due to resource constraints of training 18 individual agents. 

\textbf{Baseline Methods.} We study the effectiveness of our problem formulation by \highlight{benchmarking against two language-conditioned baselines}: \bcz and C2FARM-BC.
\highlight{\bcz is an image-to-action agent similar to BC-Z~\citep{jang2022bc}}. Following BC-Z, \highlight{we use FiLM~\citep{perez2018film} for conditioning with   CLIP~\citep{radfordLearningTransferableVisual2021} language features}, but the vision encoders take in RGB-D images instead of just RGB. We also  study both CNN and ViT vision encoders. \unet is a 3D fully-convolutional network by James et al.~\citep{c2farm} that has achieved state-of-the-art results on RLBench tasks. Similar to our agent, \unet also detects actions in a voxelized space, however it uses a coarse-to-fine-grain scheme to detect actions at two-levels of voxelization: $32^3$ voxels with a $1^3$m grid, and $32^3$ voxels with a $0.15^3$m grid after ``zooming in'' from the first level. Note that at the finest level, \unet has a higher resolution ($0.47$cm) than \model~($1$cm). We use the same 3D ConvNet architecture as James et al.~\citep{c2farm}, but instead of training it with RL, we do BC with cross-entropy loss (from \secref{sec:training}). \highlight{We also condition it with CLIP~\citep{radfordLearningTransferableVisual2021} language features at the bottleneck like in LingUNets~\citep{misra2018mapping,cliport}}. 

\input{tables/multi_task_results}
\textbf{Multi-Task Performance.} \tabref{table:rlbench} compares the performance of \bcz and \unet against \model. \highlight{With insufficient demonstrations, \bcz has near zero performance on most tasks. \bcz is disadvantaged with single-view observations and has to learn hand-eye coordination from scratch. In contrast, \model's voxel-based formulation naturally allows for integrating multi-view observations, learning 6-DoF action representations, and data-augmentation in 3D, all of which are non-trivial to achieve in image-based methods}. \unet is the most competitive baseline, but it has a limited receptive field mostly because of the coarse-to-fine-grain scheme and partly due to the convolution-only architecture. \model~outperforms \unet in $25/36$~evaluations in \tabref{table:rlbench} \highlight{with \textbf{an average improvement of} $\mathbf{1.33\times}$ \textbf{with 10 demonstrations and} $\mathbf{2.83\times}$\textbf{ with 100 demonstrations}}. For a number of tasks, \unet actually performs worse with more demonstrations, likely due to insufficient capacity. Since additional training demonstrations include additional task variants to optimize for, they might end up hurting performance. 

\vspace{-0.05cm}
In general, 10 demonstrations are sufficient for \model~to achieve $>65\%$ success on tasks with limited variations like \texttt{open drawer} (3 variations). But tasks with more variations like \texttt{stack blocks} (60 variations) need substantially more data, sometimes to simply cover all possible concepts like ``\textit{teal color block}'' that might have not appeared in the training data. See the simulation rollouts in the supplementary video to get a sense of the complexity of these evaluations. For three tasks: \texttt{insert peg}, \texttt{stack cups}, and \texttt{place cups}, all agents achieve near zero success. These are  very high-precision tasks where being off by a few centimeters or degrees could lead to unrecoverable failures. \highlight{But in \appsecref{app:high_pres} we find that training single-task agents, specifically for these tasks, slightly alleviates this issue.}

\begin{wrapfigure}{r}{0.4\textwidth}
  
  \begin{center}
    \vspace{-0.8cm}
    \includegraphics[width=0.42\textwidth]{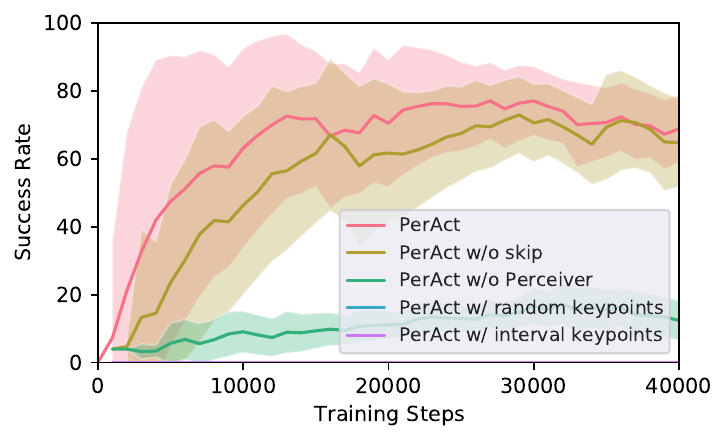}

  \vspace{-0.05cm}
  \caption{\textbf{Ablation Experiments.} Success rate of \model~after ablating key components.}
  \label{fig:ablations}
  \end{center}
  \vspace{-1em}
\end{wrapfigure}
\vspace{-0.05cm}
\textbf{Ablations.} \tabref{table:rlbench} reports \model~w/o Lang, an agent without any language conditioning. Without a language goal, the agent does not know the underlying task and performs at chance. We also report additional ablation results on the \texttt{open drawer} task in \figref{fig:ablations}. To summarize these results: (1) the skip connection helps train the agent slightly faster, (2) the Perceiver Transformer is crucial for achieving good performance with the global receptive field, and (3) extracting good keyframes actions is essential for supervised training as randomly chosen or fixed-interval keyframes lead to zero-performance.

\begin{wrapfigure}{r}{0.4\textwidth}
    \vspace{-1.63cm}
  \begin{center}
    \includegraphics[width=0.55\textwidth]{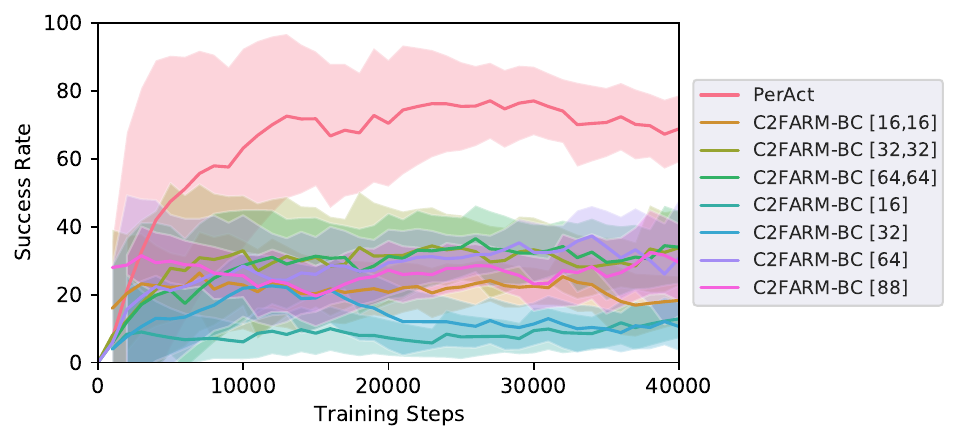}
  \end{center}
  \vspace{-1.2em}
  \caption{\textbf{Global vs. Local Receptive Field Experiments.} Success rates of \model~against various \unet~\citep{c2farm} baselines}
  \label{fig:recep_results}
  \vspace{-0.5cm}
\end{wrapfigure}
\textbf{Sensitivity Analysis.} In \appsecref{app:ablations} we investigate factors that affect \model's performance: the number of Perceiver latents, voxelization resolution, and data augmentation. We find that more latent vectors generally improve the capacity of the agent to model more tasks, but for simple short-horizon tasks, fewer latents are sufficient. Similarly, with different voxelization resolutions, some tasks are solvable with coarse voxel grids like $32^3$, but some high-precision tasks require the full $100^3$ grid. Finally, rotation perturbations in the data augmentation generally help in improving robustness essentially by exposing the agent to more rotation variations of objects.

\subsection{Global vs. Local Receptive Fields}
To further investigate our Transformer agent's global receptive field, we conduct additional experiments on the \texttt{open drawer} task. 
The \texttt{open drawer} task has three variants: \textit{``open the top drawer''}, \textit{``open the middle drawer''}, and \textit{``open the bottom drawer''}, and with a limited receptive field it is hard to distinguish the drawer handles, which are all visually identical. 
\figref{fig:recep_results} reports \model~and \unet agents trained with 100 demonstrations. Although the \texttt{open drawer} tasks can be solved with fewer demonstrations, here we want to ensure that insufficient data is not an issue. We include several versions of \unet with different voxelization schemes. For instance, $[16,16]$ indicates two levels of $16^3$ voxel grids at $1\textrm{m}^3$ and $0.15\textrm{m}^3$, respectively. And $[64]$ indicates a single level of a $64^3$ voxel grid without the coarse-to-fine-grain scheme. \model~is the only agent that achieves $>70\%$ success, whereas all \unet versions perform at chance with $\sim 33\%$, indicating that the global receptive field of the Transformer is crucial for solving the task.

\input{tables/real_robot_results}
\subsection{Real-Robot Results} \label{sec:real_robot_results}
We also validated our results with real-robot experiments on a Franka Emika Panda. See  \appsecref{app:robot_setup} for setup details. Without any sim-to-real transfer or pre-training, we trained a multi-task \model~agent \textit{from scratch }on 7 tasks (with 18 unique variations) from a total of just 53 demonstrations. See the supplementary video for qualitative results that showcase the diversity of tasks and robustness to scene changes. \tabref{table:real} reports success rates from small-scale evaluations. Similar to the simulation results, we find that \model~is able to achieve $>65\%$ success on simple short-horizon tasks like pressing hand-sanitizers from just a handful number of demonstrations. The most common failures involved predicting incorrect gripper open actions, which often lead the agent into unseen states. This could be addressed in future works by using HG-DAgger style approaches to correct the agent~\citep{jang2022bc}. Other issues included the agent exploiting biases in the dataset like in prior work~\citep{cliport}. This could be addressed by scaling up expert data with more diverse tasks and task variants.

\vspace{-0.1cm}
\section{Limitations and Conclusion}
\vspace{-0.1cm}
We presented \model, a Transformer-based multi-task agent for 6-DoF manipulation. Our experiments with both simulated and real-world tasks indicate that the right problem formulation, \ie~detecting voxel actions, makes a substantial difference in terms of data efficiency and robustness. 

While \model~is quite capable, extending it to dexterous continuous control remains a challenge. \model~is at the mercy of a sampling-based motion-planner to execute discretized actions, and is not easily extendable to N-DoF actuators like multi-fingered hands. See \appsecref{app:limitations} for an extended discussion on \model's limitations.
But overall, we are excited about scaling up robot learning with Transformers by focusing on \textit{diverse} rather than narrow multi-task data for robotic manipulation. 

\acknowledgments{We thank Selest Nashef and Karthik Desingh for their help with the Franka setup at UW. We thank Stephen James for helping with RLBench and ARM issues. We are also grateful to Valts Blukis, Zoey Chen, Markus Grotz, Aaron Walsman, and Kevin Zakka, for providing feedback on the initial draft. And thanks to Shikhar Bahl for initial discussions. This work was funded in part by ONR under award \#1140209-405780. Mohit Shridhar is supported by the NVIDIA Graduate Fellowship, and was also a part-time intern at NVIDIA throughout the duration of this project.}

\bibliography{bib}  %

\newpage
\appendix

\section{Task Details} \label{app:task_details}
\input{tables/task_desc}

\textbf{Setup.} Our simulated experiments are set in  RLBench~\citep{james2020rlbench}. We select 18 out of 100 tasks that involve at least two or more variations to evaluate the multi-task capabilities of agents. While \model~could be easily applied to more RLBench tasks, in our experiments, we were specifically interested grounding diverse language instructions, rather than learning one-off policies for single-variation tasks like \textit{``[always] take off the saucepan lid''}. 
Some tasks were modified to include additional variations. 
See \tabref{table:task_desc} for an overview. We report average keyframes extracted from the method described in \secref{sec:keyframe_and_voxel}.

\textbf{Variations.} Task variations include randomly sampled colors, sizes, shapes, counts, placements, and categories of objects. The set of colors include 20 instances: \texttt{colors} = $\{$\texttt{red}, \texttt{maroon}, \texttt{lime}, \texttt{green}, \texttt{blue}, \texttt{navy}, \texttt{yellow}, \texttt{cyan}, \texttt{magenta}, \texttt{silver}, \texttt{gray}, \texttt{orange}, \texttt{olive}, \texttt{purple}, \texttt{teal}, \texttt{azure}, \texttt{violet}, \texttt{rose}, \texttt{black}, \texttt{white}$\}$. The set of sizes include 2 instances: \texttt{sizes} = $\{$\texttt{short}, \texttt{tall}$\}$. The set of shapes include 5 instances: \texttt{shapes} = $\{$\texttt{cube}, \texttt{cylinder}, \texttt{triangle}, \texttt{star}, \texttt{moon}$\}$. The set of counts include 3 instances: \texttt{counts} = $\{$\texttt{1}, \texttt{2}, \texttt{3}$\}$. The placements and object categories are specific to each task. For instance, \texttt{open drawer} has 3 placement locations: \texttt{top}, \texttt{middle}, and \texttt{bottom}, and \texttt{put in cupboard} includes 9 YCB objects. In addition to these semantic variations, objects are placed on the tabletop at random poses. Some large objects like drawers have constrained pose variations~\citep{james2020rlbench} to ensure that manipulating them is kinematically feasible with the Franka arm.
 
In the following sections, we describe each of 18 tasks in detail. We highlight tasks that were modified from the original RLBench~\citep{james2020rlbench} codebase\footnote{\url{https://github.com/stepjam/RLBench}} and describe what exactly was modified.

\subsection{Open Drawer}
\textbf{Filename:} $\texttt{open\_drawer.py}$ 

\textbf{Task:} Open one of the three drawers: \texttt{top}, \texttt{middle}, or \texttt{bottom}.

\textbf{Modified:} No.

\textbf{Objects:} 1 drawer.

\textbf{Success Metric}: The prismatic joint of the specified drawer is fully extended. 

\subsection{Slide Block}
\textbf{Filename:} $\texttt{slide\_block\_to\_color\_target.py}$ 

\textbf{Task:} Slide the block on to one of the colored square targets. The target colors are limited to $\texttt{red}$, $\texttt{blue}$, $\texttt{pink}$, and $\texttt{yellow}$.

\textbf{Modified:} Yes. The original $\texttt{slide\_block\_to\_target.py}$ task contained only one target. Three other targets were added to make a total of 4 variations.

\textbf{Objects:} 1 block and 4 colored target squares.

\textbf{Success Metric}: Some part of the block is inside the specified target area.

\subsection{Sweep to Dustpan}
\textbf{Filename:} $\texttt{sweep\_to\_dustpan\_of\_size.py}$ 

\textbf{Task:} Sweep the dirt particles to either the short or tall dustpan.

\textbf{Modified:} Yes. The original $\texttt{sweep\_to\_dustpan.py}$ task contained only one dustpan. One other dustpan was added to make a total of 2 variations. 

\textbf{Objects:} 5 dirt particles and 2 dustpans.

\textbf{Success Metric}: All 5 dirt particles are inside the specified dustpan. 

\subsection{Meat Off Grill}
\textbf{Filename:} $\texttt{meat\_off\_grill.py}$ 

\textbf{Task:} Take either the chicken or steak off the grill and put it on the side. 

\textbf{Modified:} No.

\textbf{Objects:} 1 piece of chicken, 1 piece of steak, and 1 grill.

\textbf{Success Metric}: The specified meat is on the side, away from the grill. 

\subsection{Turn Tap}
\textbf{Filename:} $\texttt{turn\_tap.py}$ 

\textbf{Task:} Turn either the left or right handle of the tap. Left and right are defined with respect to the faucet orientation.  

\textbf{Modified:} No.

\textbf{Objects:} 1 faucet with 2 handles.

\textbf{Success Metric}: The revolute joint of the specified handle is at least $90^\circ$ off from the starting position. 

\subsection{Put in Drawer}
\textbf{Filename:} $\texttt{put\_item\_in\_drawer.py}$ 

\textbf{Task:} Put the block in one of the three drawers: \texttt{top}, \texttt{middle}, or \texttt{bottom}.

\textbf{Modified:} No.

\textbf{Objects:} 1 block and 1 drawer.

\textbf{Success Metric}: The block is inside the specified drawer.

\subsection{Close Jar}
\textbf{Filename:} $\texttt{close\_jar.py}$ 

\textbf{Task:} Put the lid on the jar with the specified color and screw the lid in. The jar colors are sampled from the full set of 20 color instances.

\textbf{Modified:} No.

\textbf{Objects:} 1 block and 2 colored jars. 

\textbf{Success Metric}: The lid is on top of the specified jar and the Franka gripper is not grasping anything.

\subsection{Drag Stick}
\textbf{Filename:} $\texttt{reach\_and\_drag.py}$ 

\textbf{Task:} Grab the stick and use it to drag the cube on to the specified colored target square. The target colors are sampled from the full set of 20 color instances. 

\textbf{Modified:} Yes. The original $\texttt{reach\_and\_drag.py}$ task contained only one target. Three other targets were added with randomized colors.

\textbf{Objects:} 1 block, 1 stick, and 4 colored target squares.

\textbf{Success Metric}: Some part of the block is inside the specified target area.

\subsection{Stack Blocks}
\textbf{Filename:} $\texttt{stack\_blocks.py}$ 

\textbf{Task:} Stack $N$ blocks of the specified color on the green platform. There are always 4 blocks of the specified color, and 4 distractor blocks of another color. The block colors are sampled from the full set of 20 color instances. 

\textbf{Modified:} No.

\textbf{Objects:} 8 color blocks (4 are distractors), and 1 green platform.

\textbf{Success Metric}: $N$ blocks are inside the area of the green platform. 

\subsection{Screw Bulb}
\textbf{Filename:} $\texttt{light\_bulb\_in.py}$ 

\textbf{Task:} Pick up the light bulb from the specified holder, and screw it into the lamp stand. The colors of holder are sampled from the full set of 20 color instances. There are always two holders in the scene -- one specified and one distractor holder. 

\textbf{Modified:} No.

\textbf{Objects:} 2 light bulbs, 2 holders, and 1 lamp stand.

\textbf{Success Metric}: The bulb from the specified holder is inside the lamp stand dock. 

\subsection{Put in Safe}
\textbf{Filename:} $\texttt{put\_money\_in\_safe.py}$ 

\textbf{Task:} Pick up the stack of money and put it inside the safe on the specified shelf. The shelf has three placement locations: \texttt{top}, \texttt{middle}, \texttt{bottom}. 

\textbf{Modified:} No.

\textbf{Objects:} 1 stack of money, and 1 safe. 

\textbf{Success Metric}: The stack of money is on the specified shelf inside the safe. 

\subsection{Place Wine}
\textbf{Filename:} $\texttt{place\_wine\_at\_rack\_location.py}$ 

\textbf{Task:} Grab the wine bottle and put it on the wooden rack at one of the three specified locations: \texttt{left}, \texttt{middle}, \texttt{right}. The locations are defined with respect to the orientation of the wooden rack. 

\textbf{Modified:} Yes. The original \texttt{stack\_wine.py} task had only one placement location. Two other locations were added to make a total of 3 variations. 

\textbf{Objects:} 1 wine bottle, and 1 wooden rack.

\textbf{Success Metric}: The wine bottle is at the specified placement location on the wooden rack. 

\subsection{Put in Cupboard}
\textbf{Filename:} $\texttt{put\_groceries\_in\_cupboard.py}$ 

\textbf{Task:} Grab the specified object and put it in the cupboard above. The scene always contains 9 YCB objects that are randomly placed on the tabletop.  

\textbf{Modified:} No.

\textbf{Objects:} 9 YCB objects, and 1 cupboard (that hovers in the air like magic). 

\textbf{Success Metric}: The specified object is inside the cupboard.

\subsection{Sort Shape}
\textbf{Filename:} $\texttt{place\_shape\_in\_shape\_sorter.py}$ 

\textbf{Task:} Pick up the specified shape and place it inside the correct hole in the sorter. There are always 4 distractor shapes, and 1 correct shape in the scene.

\textbf{Modified:} Yes. The sizes of the shapes and sorter were enlarged so that they are distinguishable in the RGB-D input. 

\textbf{Objects:} 5 shapes, and 1 sorter.

\textbf{Success Metric}: The specified shape is inside the sorter.

\subsection{Push Buttons}
\textbf{Filename:} $\texttt{push\_buttons.py}$ 

\textbf{Task:} Push the colored buttons in the specified sequence. The button colors are sampled from the full set of 20 color instances. There are always three buttons in scene. 

\textbf{Modified:} No.

\textbf{Objects:} 3 buttons.

\textbf{Success Metric}: All the specified buttons were pressed.

\subsection{Insert Peg}
\textbf{Filename:} $\texttt{insert\_onto\_square\_peg.py}$ 

\textbf{Task:} Pick up the square and put it on the specified color spoke. The spoke colors are sampled from the full set of 20 color instances. 

\textbf{Modified:} No.

\textbf{Objects:} 1 square, and 1 spoke platform with three color spokes.

\textbf{Success Metric}: The square is on the specified spoke.

\subsection{Stack Cups}
\textbf{Filename:} $\texttt{stack\_cups.py}$ 

\textbf{Task:} Stack all cups on top of the specified color cup. The cup colors are sampled from the full set of 20 color instances. The scene always contains three cups.   

\textbf{Modified:} No.

\textbf{Objects:} 3 tall cups.

\textbf{Success Metric}: All other cups are inside the specified cup.

\subsection{Place Cups}
\textbf{Filename:} $\texttt{place\_cups.py}$ 

\textbf{Task:} Place $N$ cups on the cup holder. This is a very high precision task where the handle of the cup has to be exactly aligned with the spoke of the cup holder for the placement to succeed. 

\textbf{Modified:} No.

\textbf{Objects:} 3 cups with handles, and 1 cup holder with three spokes.

\textbf{Success Metric}: $N$ cups are on the cup holder, each on a separate spoke.

\section{\model~Details} \label{app:peract_details}
In this section, we provide implementation details for \model. See this \href{https://colab.research.google.com/drive/1wpaosDS94S0rmtGmdnP0J1TjS7mEM14V?usp=sharing}{Colab tutorial} for a PyTorch implementation.

\textbf{Input Observation.} Following James et al.~\citep{c2farm}, our input voxel observation is a $100^3$ voxel grid with $10$ channels:  $\mathbb{R}^{100 \times 100 \times 100 \times 10}$. The grid is constructed by fusing calibrated pointclouds with PyTorch's \texttt{scatter\_}  function\footnote{\url{https://pytorch.org/docs/stable/generated/torch.Tensor.scatter_.html}}. The $10$ channels are composed of: $3$ RGB, $3$ point, $1$ occupancy, and $3$ position index values. The RGB values are normalized to a zero-mean distribution. The point values are Cartesian coordinates in the robot's coordinate frame. The occupancy value indicates if a voxel is occupied or empty. The  position index values represent the 3D location of the voxel with respect to the $100^3$ grid. In addition to the voxel observation, the input also includes proprioception data with $4$ scalar values: gripper open, left finger joint position, right finger joint position, and timestep (of the action sequence). 

\textbf{Input Language.} The language goals are encoded with CLIP's language encoder~\citep{radfordLearningTransferableVisual2021}. We use CLIP's tokenizer to preprocess the sentence, which always results in an input sequence of $77$ tokens (with zero-padding). These tokens are encoded with the language encoder to produce a sequence of dimensions $\mathbb{R}^{77 \times 512}$. 

\textbf{Preprocessing.} The voxel grid is encoded with a 3D convolution layer with a $1\times1$ kernel to upsample the channel dimension from $10$ to $64$. Similarly, the proprioception data is encoded with a linear layer to upsample the input dimension from $4$ to $64$. The encoded voxel grid is split into $5^3$ patches through a 3D convolution layer with a kernel-size and stride of 5, which results in a patch tensor of dimensions $\mathbb{R}^{20 \times 20 \times 20 \times 64}$. The proprioception features are tiled in 3D to match the dimensions of the patch tensor, and concattenated along the channel to form a tensor of dimensions $\mathbb{R}^{20 \times 20 \times 20 \times 128}$. This tensor is flattened into a sequence of dimensions $\mathbb{R}^{8000 \times 128}$. The language features are downsampled with a linear layer from $512$ to $128$ dimensions, and then appended to the tensor to form the final input sequence to the Perceiver Transformer, which of dimensions $\mathbb{R}^{8077 \times 128}$. We also add learned positional embeddings to the input sequence. These embeddings are represented with trainable \texttt{nn.Parameter}(s) in PyTorch.  

\begin{wrapfigure}[13]{r}{0.6\textwidth}
  \vspace{-0.7cm}
  \begin{center}
    \includegraphics[width=0.65\textwidth]{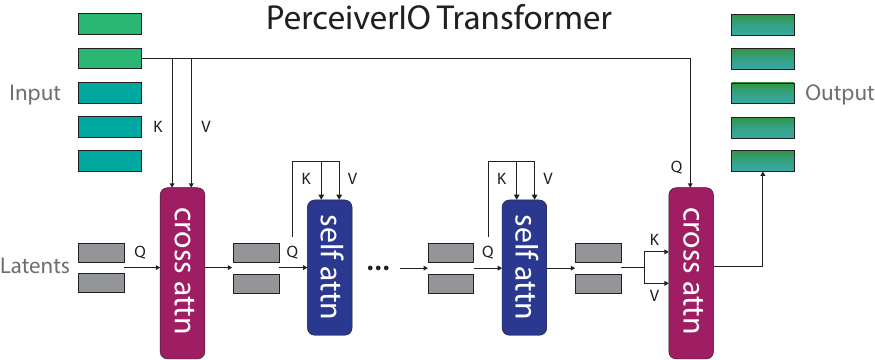}
  \caption{\textbf{Perceiver Transformer Architecture.} Perceiver is a latent-space transformer. Q, K, V represent queries, keys, and values, respectively. We use 6 self-attention layers in our implementation.}
  \vspace{2cm}
  \label{fig:perceiver_arch}
  \end{center}
\end{wrapfigure}
\textbf{Perceiver Transformer} is a latent-space Transformer~\citep{jaegle2021perceiver} that uses a small set of latent vectors to encode extremely long input sequences. See \figref{fig:perceiver_arch} for an illustration of this process. 
Perceiver first computes cross-attention between the input sequence and the set of latent vectors of dimensions $\mathbb{R}^{2048 \times 512}$. 
These latents are randomly initialized and trained end-to-end. The latents are encoded with 6 self-attention layers, and then cross-attended with the input to output a sequence that matches the input-dimensions. This output is upsampled with a 3D convolution layer and tri-linear upsampling to form a voxel feature grid with $64$ channels: $\mathbb{R}^{100 \times 100 \times 100 \times 64}$. 
This feature grid is concatenated with the initial $64$-dimensional feature grid from the processing stage as a skip connection to the encoding layers. 
Finally, a 3D convolution layer with a $1 \times 1$ kernel downsamples the channels from $128$ back to $64$ dimensions.
Our implementation of Perceiver is based on an existing open-source repository\footnote{\url{https://github.com/lucidrains/perceiver-pytorch}}.

\textbf{Decoding.} For translation, the voxel feature grid is decoded with a 3D convolution layer with a $1\times1$ kernel to downsample the channel dimension from $64$ to $1$. This tensor is the translation $\mathcal{Q}$-function of dimensions $\mathbb{R}^{100 \times 100 \times 100 \times 1}$. For rotation, gripper open, and collision avoidance actions, the voxel feature grid is max-pooled along the 3D dimensions to form a vector of dimensions $\mathbb{R}^{1 \times 64}$. This vector is decoded with three independent linear layers to form the respective $\mathcal{Q}$-functions for rotation, gripper open, and collision avoidance. The rotation linear layer outputs logits of dimensions $\mathbb{R}^{216}$ ($72$ bins of 5 degree increments for each of the three axes). The gripper open and collide linear layers output logits of dimensions $\mathbb{R}^{2}$.

Our codebase is built on the ARM repository\footnote{\url{https://github.com/stepjam/ARM}} by James et al.~\citep{c2farm}.  

\section{Evaluation Workflow}

\subsection{Simulation}
Simulated experiments in \secref{sec:sim_results} follow a four-phase workflow: (1) generate a dataset with train, validation, and test sets, each containing $100$, $25$, and $25$ demonstrations, respectively. (2) Train an agent on the train set and save checkpoints at intervals of 10K iterations. (3) Evaluate all saved checkpoints on the validation set, and mark the best performing checkpoint. (4) Evaluate the best performing checkpoint on the test set. While this workflow follows a standard train-val-test paradigm from supervised learning, it is not the most feasible workflow for real-robot settings. With real-robots, collecting a validation set and evaluating all checkpoints could be very expensive.  

\subsection{Real-Robot}

For real-robot experiments in \secref{sec:real_robot_results}, we simply pick the last checkpoint from training. We check if the agent has been sufficiently trained by visualizing $\mathcal{Q}$-predictions on training examples with swapped or modified language goals. While evaluating a trained agent, the agent keeps acting until a human user stops the execution. We also visualize the $\mathcal{Q}$-predictions live to ensure that the agent's upcoming action is safe to execute.

\newpage
\section{Robot Setup} \label{app:robot_setup}

\subsection{Simulation}

\begin{wrapfigure}[10]{r}{0.5\textwidth}
  \vspace{-0.60cm}
  \begin{center}
    \includegraphics[width=0.45\textwidth]{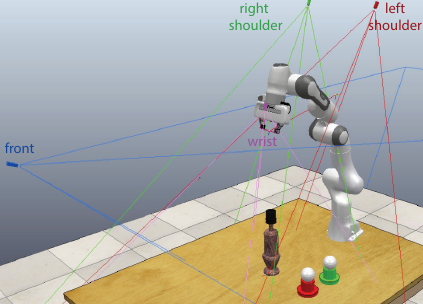}
  \caption{\textbf{Simulated Setup.} The four camera setup: front, left shoulder, right shoulder, and on the wrist.}
  \label{fig:sim_setup}
  \end{center}
  \vspace{-1cm}
\end{wrapfigure}

All simulated experiments use the four camera setup illustrated in \figref{fig:sim_setup}. The front, left shoulder, and right shoulder cameras, are static, but the wrist camera moves with the end-effector. We did not modify the default camera poses from RLBench~\citep{james2020rlbench}. These poses maximize coverage of the tabletop, while minimizing occlusions caused by the moving arm. The wrist camera in particular is able to provide high-resolution observations of small objects like handles. 

\vspace{0.6cm}
\subsection{Real-Robot}

\textbf{Hardware Setup}. The real-robot experiments use a Franka Panda manipulator with a parallel-gripper. For perception, we use a Kinect-2 RGB-D camera mounted on a tripod, at an angle, pointing towards the tabletop. See \figref{app:robot_setup} for reference. We tried setting-up multiple Kinects for multi-view observations, but we could not fix the interference issue caused by multiple Time-of-Flight sensors. The Kinect-2 provides RGB-D images of resolution $512 \times 424$ at 30Hz. The extrinsics between the camera and robot base-frame are calibrated with the \texttt{easy\_handeye}  package\footnote{\url{https://github.com/IFL-CAMP/easy_handeye}}. We use an ARUCO\footnote{\url{https://github.com/pal-robotics/aruco_ros}} AR marker mounted on the gripper to aid the calibration process. 

\begin{wrapfigure}[20]{r}{0.5\textwidth}
  \vspace{-0.7cm}
  \begin{center}
    \includegraphics[width=0.5\textwidth]{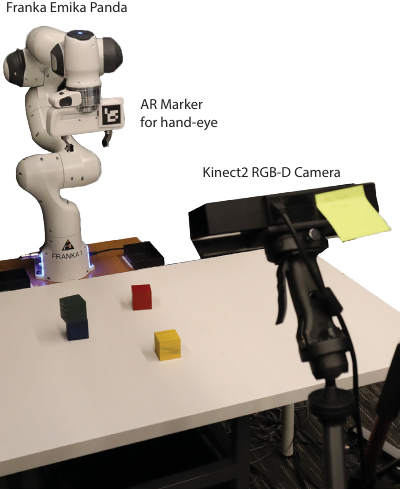}
  \caption{\textbf{Real-Robot Setup} with Kinect-2 and Franka Panda.}
  \label{fig:robot_setup}
  \end{center}
\end{wrapfigure}

\textbf{Data Collection.} We collect demonstrations with an HTC Vive controller. The controller is a 6-DoF tracker that provides accurate poses with respect to a static base-station. These poses are displayed as a marker on RViz\footnote{\url{http://wiki.ros.org/rviz}} along with the real-time RGB-D pointcloud from the Kinect-2. A user specifies target poses by using the marker and pointcloud as reference. These target poses are executed with a motion-planner. We use Franka ROS and MoveIt\footnote{\url{http://docs.ros.org/en/kinetic/api/moveit_tutorials/html/}}, which by default uses an RRT-Connect planner. 

\textbf{Training and Execution.} We train a \model~agent from scratch with 53 demonstrations. The training samples are augmented with $\pm0.125$m translation perturbations and $\pm45^\circ$ yaw rotation perturbations. We train on 8 NVIDIA P100 GPUs for 2 days. During evaluation, we simply chose the last checkpoint from training (since we did not collect a validation set for optimization). Inference is   done on a single Titan X GPU.

\newpage
\vspace{-1cm}
\section{Data Augmentation} \label{app:data_aug}
\model's voxel-based formulation naturally allows for data augmentation with SE(3) transformations. During training, samples of voxelized observations $\mathbf{v}$ and their corresponding keyframe actions $\mathbf{k}$ are perturbed with random translations and rotations. Translation perturbations have a range of $[\pm0.125\textrm{m}, \pm0.125\textrm{m}, \pm0.125\textrm{m}]$. Rotation perturbations are limited to the yaw axis and have a range of $[0^{\circ}, 0^{\circ}, \pm45^{\circ}]$. The $45^{\circ}$ limit ensures that the perturbed rotations do not go beyond what is kinematically reachable for the Franka arm. We did experiment with pitch and roll perturbations, but they substantially lengthened the training time. Any perturbation that pushed the discretized action outside the observation voxel grid was discarded. See the bottom row of \figref{fig:qpred_extra} for examples of data augmentation.

\section{Demo Augmentation} \label{app:keypoints_and_demo}
\begin{wrapfigure}[7]{r}{0.3\textwidth}
  \vspace{-2cm}
  \begin{center}
  \hspace{1cm}
    \includegraphics[width=0.3\textwidth]{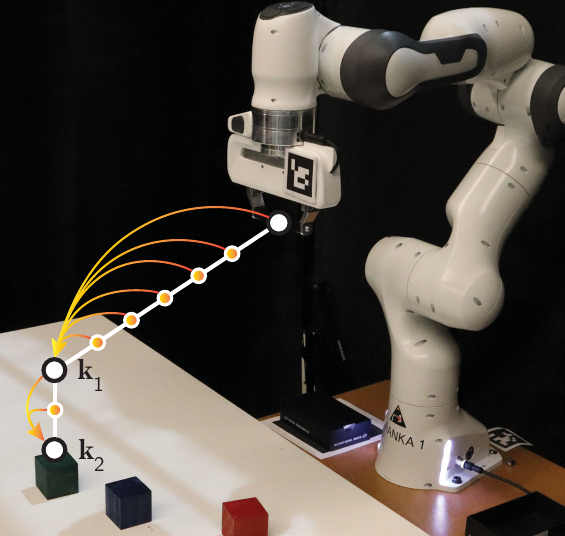}
  \caption{\textbf{Keyframes and Demo Augmentation.}}
  \label{fig:keypoints_and_demo}
  \end{center}
  \vspace{-12cm}
\end{wrapfigure}
Following James et al.~\citep{james2020rlbench}, we cast every datapoint in a demonstration as a ``predict the next (best) keyframe action'' task. See \figref{fig:keypoints_and_demo} for an illustration of this process. In this illustration, $\mathbf{k}_{1}$ and $\mathbf{k}_{2}$ are two keyframes that were extracted from the method described in \secref{sec:keyframe_and_voxel}. The orange circles indicate datapoints whose RGB-D observations are paired with the next keyframe action.

\section{Sensitivity Analysis} \label{app:ablations}
\input{tables/sen_analysis}

In \tabref{table:sen_analysis}, we investigate three factors that affect \model's performance: rotation data augmentation, number of Perceiver latents, and voxelization resolution. All multi-task agents were trained with 100 demonstrations per task and evaluated on 25 episodes per task. To briefly summarize these results: (1) $45^{\circ}$ yaw perturbations improve performance on tasks with lots of rotation variations like \texttt{stack blocks}, but also worsen performance on tasks with constrained rotations like \texttt{place wine}. (2) \model~with just $512$ latents is competitive with (and sometimes even better than) the default agent with $2048$ latents, which showcases the compression capability of the Perceiver architecture. (3) Coarse grids like $32^3$ are sufficient for some tasks, but  high-precision tasks like \texttt{sort shape} need higher resolution voxelization. (4) Large patch-sizes reduce memory usage, but they might affect tasks that need sub-patch precision.

\section{High-Precision Tasks}
\label{app:high_pres}

\input{tables/high_pres_res}
In \tabref{table:rlbench}, \model~achieves zero performance on three high-precision tasks: \texttt{place cups}, \texttt{stack cups}, and \texttt{insert peg}. To investigate if multi-task optimization is itself one of the factors affecting performance, we train 3 separate single-task agents for each task. We find that single-task agents are able to achieve non-zero performance, indicating that better multi-task optimization methods might  improve performance on certain tasks. 

\section{Additional Related Work}

In this section, we briefly discuss additional works that were not mentioned in \secref{sec:related_work}.

\highlight{\textbf{Concurrent Work.} Recently, Mandi et al.~\citep{mandi2022effectiveness} found that pre-training and fine-tuning on new tasks is competitive, or even better, than meta-learning approaches for RLBench tasks in multi-task (but single-variation) settings. This pre-training and fine-tuning paradigm might be directly applicable to~\model, where a pre-trained \model ~agent could be quickly adapted to new tasks without the explicit use of meta-learning algorithms.}

\highlight{\textbf{Multi-Task Learning.} In the context of RLBench, Auto-$\lambda$~\citep{liu2022auto_lambda} presents a multi-task optimization framework that goes beyond uniform task weighting from \secref{sec:training}. The method dynamically tunes task weights based on the validation loss. Future works with \model~could replace uniform task weighting with Auto-$\lambda$ for better multi-task performance. In the context of Meta-World~\citep{yu2020meta}, Sodhani et al.~\citep{pmlr-v139-sodhani21a} found that language-conditioning leads to performance gains for multi-task RL on 50 task variations.}

\highlight{\textbf{Language-based Planning.} In this paper, we only investigated single-goal settings where the language instruction does not change throughout the episode. However, language-conditioning natural allows for composing several instructions in a sequential manner~\citep{lynch2020grounding}. As such, several prior  works~\citep{ALFWorld20,ahn2022can,zeng2022socratic,huang2022language} have used language as medium for planning high-level actions, which can then be executed with pre-trained low-level skills. Future works could incorporate language-based planning for grounding more abstract goals like \textit{``make dinner''}.}

\textbf{Task and Motion Planning.} In the sub-field of Task and Motion Planning (TAMP)~\citep{kaelbling2013integrated,garrett2021integrated}, Konidaris \etal~\citep{konidaris2018skills} present an action-centric approach to symbolic planning. Given a set of predefined action-skills, an agent interacts with its environment to construct a set of symbols, which can then be used for planning.

\textbf{Voxel Representations.} Voxel-based representations have been used in several domains that specifically benefit from 3D understanding.
Like in object detection~\citep{mao2021voxel,he2022voxel}, object search~\citep{zheng2022towards}, and vision-language grounding~\citep{blukis2022persistent,corona-etal-2022-voxel}, voxel maps have been used to build persistent scene representations~\citep{sitzmann2019deepvoxels}.
In Neural Radiance Fields (NeRFs), voxel feature grids have dramatically reduced training and rendering times~\citep{mueller2022instant,yu_and_fridovichkeil2021plenoxels}. Similarly, other works in robotics have used voxelized representations to embed viewpoint-invariance for driving~\citep{lal2021coconets} and manipulation~\citep{tung20203d}. The use of latent vectors in Perceiver~\citep{jaegle2021perceiver} is broadly related to voxel hashing~\citep{niessner2013real} from computer graphics. Instead of using a location-based hashing function to map voxels to fixed size memory, PerceiverIO uses cross attention to map the input to fixed size latent vectors, which are trained end-to-end. Another major difference is the treatment of unoccupied space. In graphics, unoccupied space does not affect  rendering, but in \model, unoccupied space is where a lot of ``action detections'' happen. Thus the relationship between unoccupied and occupied space, \ie scene, objects, robot, is crucial for learning action representations.

\textbf{Long-Context and Latent-Space Transformers.} Several approaches have been proposed for extending Transformers to longer context lengths~\citep{tay2020efficient}. Latent-space Transformers that use fixed-size latents instead of the full context, are one such approach~\citep{jaegle2021perceiver,goyal2021coordination}. There is no clear winner in terms of trade-offs between speed, memory, and performance. However, latent-space methods have achieved compelling results in object detection~\citep{carion2020end} and slot-attention based object discovery~\citep{locatello2020object}.

\section{Additional Q-Prediction Examples} \label{app:more_qpred}
\begin{figure*}[h]
    \centering
    \hspace*{-3.1cm}
    \includegraphics[width=1.4\textwidth]{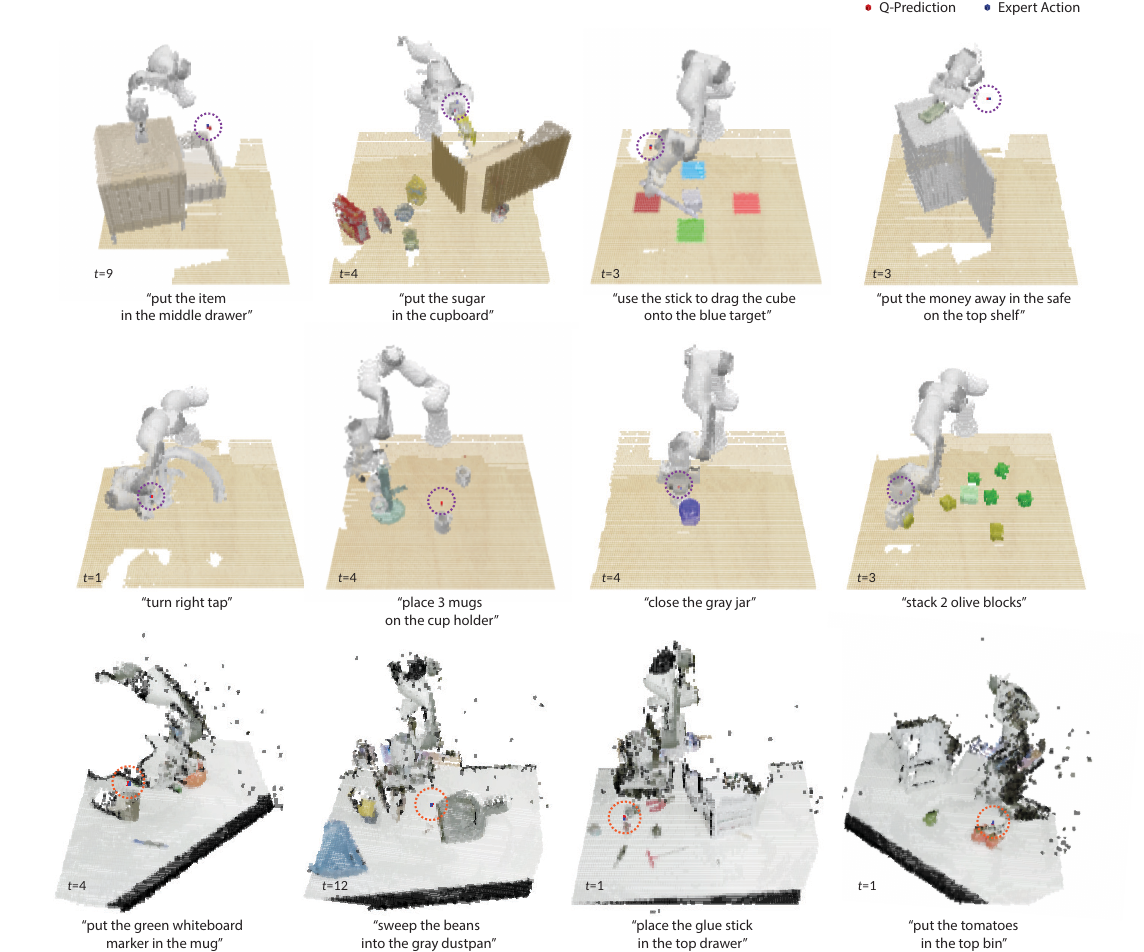}
    \caption{\textbf{Additional Q-Prediction Examples.} Translation $\mathcal{Q}$-Prediction examples from \model. The top two rows are from simulated tasks without any data augmentation perturbations, and the bottom row is from real-world tasks  with translation and yaw-rotation perturbations.}
    \label{fig:qpred_extra}
\end{figure*}

\figref{fig:qpred_extra} showcases additional $\mathcal{Q}$-prediction examples from trained \model~agents. Traditional object-centric representations like poses and instance-segmentations struggle to represent piles of beans or tomato vines with high-precision. Whereas action-centric agents like \model~focus on learning perceptual representations of actions, which elevates the need for practitioners to define \textit{what should be an object} (which is a harder problem and often specific to tasks and embodiments). 

\newpage
\section{Things that did not work}
\vspace{-0.2cm}
In this section, we describe things we tried, but did not work or caused issues in practice. 

\textbf{Real-world multi-camera setup.} We tried setting up multiple Kinect-2s for real-world multi-view observations, but we could not solve interference issues with multiple Time-of-Flight sensors. Particularly, the depth frames became very noisy and had lots of holes. Future works could try turning the cameras on-and-off in a rapid sequence, or use better Time-of-Flight cameras with minimal interference. 

\textbf{Fourier features for positional embeddings.} Instead of the learned positional embeddings, we also experimented with concatenating Fourier features to the input sequence like in some Perceiver models~\citep{jaegle2021perceiver}. The Fourier features led to substantially worse performance. 

\textbf{Pre-trained vision features.} Following CLIPort~\citep{cliport}, we tried using pre-trained vision features from CLIP~\citep{radfordLearningTransferableVisual2021}, instead of raw RGB values, to bootstrap learning and also to improve generalization to unseen objects. We ran CLIP's ResNet50 on each of the 4 RGB frames, and upsampled features with shared  decoder layers in a UNet fashion. But we found this to be extremely slow, especially since the ResNet50 and decoder layers need to be run on 4 independent RGB frames. With this additional overhead, training multi-task agents would have taken substantially longer than 16 days. Future works could experiment with methods for pre-training the decoder layers on auxiliary tasks, and pre-extracting features for faster training. 

\textbf{Upsampling at multiple self-attention layers.} Inspired by Dense Prediction Transformers (DPT)~\citep{ranftl2021vision}, we tried upsampling features at multiple self-attention layers in the Perceiver Transformer. But this did not work at all; perhaps the latent-space self-attention layers of Perceiver are substantially different to the full-input self-attention layers of ViT~\citep{dosovitskiy2020image} and DPT~\citep{ranftl2021vision}.

\textbf{Extreme rotation augmentation.} In addition to yaw rotation perturbations, we also tried perturbing the pitch and roll. While \model~was still able to learn policies, it took substantially longer to train. It is also unclear if the default latent size of $\mathbb{R}^{2048 \times 512}$ is appropriate for learning 6-DoF polices with such extreme rotation perturbations. 

\textbf{Using Adam instead of LAMB.} We tried training \model~with the Adam~\citep{kingma2014adam} optimizer instead of LAMB~\citep{you2019large}, but this led to worse performance in both simulated and real-world experiments. 
\vspace{-0.1cm}
\section{Limitations and Risks} \label{app:limitations}
\vspace{-0.1cm}
While \model~is quite capable, it is not without limitations. In the following sections, we discuss some of these limitations and potential risks for real-world deployment. 

\textbf{Sampling-Based Motion Planner.} \model~relies on a sampling-based motion planner to execute discretized actions. This puts \model~at the mercy of randomized planner to reach poses. While this issue did not cause any major problems with the tasks in our experiments, a lot of other tasks are sensitive to the paths taken to reach poses. For instance, pouring water into a cup would require a smooth path for tilting the water container appropriately. This could be addressed in future works by using a combination of learned and sampled motion paths~\citep{james2022coarse}.

\textbf{Dynamic Manipulation.} Another issue with discrete-time discretized actions is that they are not easily applicable to dynamic tasks that require real-time closed-loop maneuvering. This could be addressed with a separate visuo-servoing mechanism that can reach target poses with closed-loop control. Alternatively, instead of predicting just one action, \model~could be extended to predict a sequence of discretized actions. Here, the Transformer-based architecture could be particularly advantageous. Also, instead of just predicting poses, the agent could also be trained to predict other physical parameters like target velocities~\citep{zeng2020tossingbot}.

\textbf{Dexterous Manipulation.} Using discretized actions with N-DoF robots like multi-fingered hands is also non-trivial. Specifically for multi-fingered hands, \model~could be modified to predict finger-tip poses that can be reached with an IK (Inverse Kinematics) solver. But it is unclear how feasible or robust such an approach would be with under-actuated systems like multi-fingered hands.

\begin{figure*}[!t]
    \centering
    \vspace{-1.2cm}
    \includegraphics[width=0.8\textwidth]{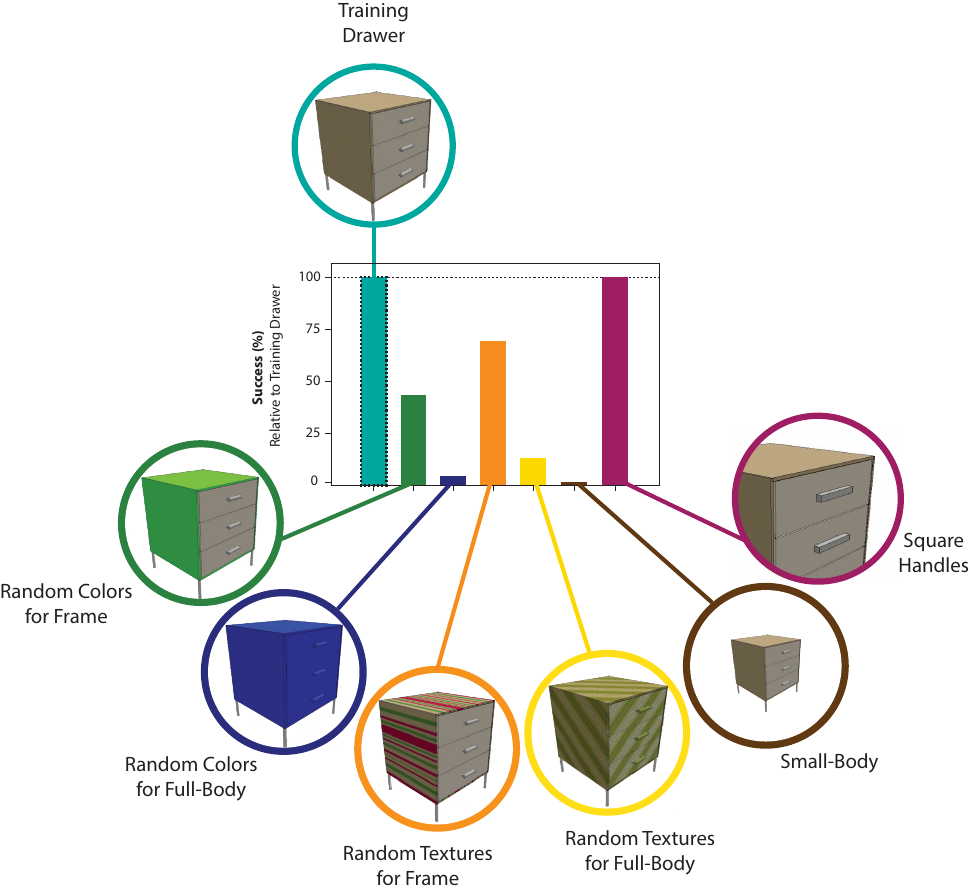}
    \caption{\textbf{Perturbation Tests.} Results from a multi-task \model~agent trained on a single drawer and evaluated on several instances perturbed drawers. Each perturbation consists of 25 evaluation episodes, and reported successes are relative to the training drawer.}
    \label{fig:robustness}
    \vspace{-1.5em}
\end{figure*}
\textbf{Generlization to Novel Instances and Objects.} 
\highlight{In \figref{fig:robustness}, we report results from small-scale perturbation experiments on the \texttt{open drawer} task. We observe that changing the shape of the handles does not affect performance. However, handles with randomized textures and colors confuse the agent since it has only seen one type of drawer color and texture during training.
Going beyond this one-shot setting, and training on several instances of drawers might improve generalization performance.}
Although we did not explicitly study generalization to unseen objects, it might be feasible to train \model's action-detector on a broad range of objects and evaluate its ability to handle novel objects, akin to how language-conditioned instance-segmentors and object-detectors are used~\citep{kamath2021mdetr}.
Alternatively, pre-trained vision features from multi-modal encoders like CLIP~\citep{radfordLearningTransferableVisual2021} or R3M~\citep{nair2022r3m} could be used to boostrap learning. 

\textbf{Scope of Language Grounding.} Like with prior work~\citep{cliport}, \model's understanding of verb-noun phrases is closely grounded in demonstrations and tasks. For example, ``cleaning'' in \textit{``clean the beans on the table with a dustpan''} is specifically associated with the action sequence of pushing beans on to a dustpan, and not ``cleaning'' in general, which could be applied to other tasks like cleaning the table with a cloth.

\textbf{Predicting Task Completion.} For both real-world and simulated evaluations, an oracle indicates whether the desired goal has been reached. This oracle could be replaced with a success classifier that can be pre-trained to predict task completion from RGB-D observations. 

\textbf{History and Partial Observability.} \model~relies purely on the current observation to predict the next action. As such, tasks that require history like counting or ordering are not feasible, unless accompanied by a task-completion predictor. Similarly, for tasks involving partial observability \eg~looking through drawers one-by-one for a specific object, \model~does not keep  track of what was seen before. Future works could include observations from previous timesteps, or append Perceiver latents, or train a Recurrent Neural Network to encode latents across timesteps.

\textbf{Data Augmentation with Kinematic Feasibility.} The data augmentation method described in \secref{app:data_aug} does not consider the kinematic feasibility of reaching perturbed actions with the Franka arm. Future works could pre-compute unreachable poses in the discretized action space, and discard any augmentation perturbations that push actions into unreachable zones.  

\textbf{Balanced Datasets.} Since \model~is trained with just a few demonstrations, it occassionally tends to exploit biases in the training data. For instance, \model~might have a tendency to always \textit{``place blue blocks on yellow blocks''} if such an example is over-represented in the training data. Such issues could be potentially fixed by scaling datasets to include more diverse examples of objects and attributes. Additionally, data visualization methods could be used to identify and fix these biases. 

\highlight{\textbf{Multi-Task Optimization.} The uniform  task sampling strategy presented in \secref{sec:training} might sometimes hurt performance. Since all tasks are weighted equally, optimizing for certain tasks with common elements (\eg~moving blocks), might adversarial affect the performance on other dissimilar tasks (\eg~turning taps). Future works, could use dynamic task-weighting methods like Auto-$\lambda$~\citep{liu2022auto_lambda} for better multi-task optimization.}

\textbf{Deployment Risks.} \model~is an end-to-end framework for 6-DoF manipulation. Unlike some methods in Task-and-Motion-Planning  that can sometimes provide theoretical guarantees on task completion, \model~is a purely reactive system whose performance can only be evaluated through empirical means. Also, unlike prior works~\citep{cliport}, we do not use internet pre-trained vision encoders that might contain harmful biases~\citep{birhane2021multimodal,bender2021dangers}. Even so, it is prudent to thoroughly study and mitigate any biases before deployment. As such, for real-world applications, keeping humans in the loop both during training and testing, might help. Usage with unseen objects and observations with people is not recommended for safety critical systems.

\section{Emergent Properties}

In this section, we present some preliminary  findings on the emergent properties of \model.

\begin{wrapfigure}{r}{0.40\textwidth}
  \vspace{-1.3cm}
  \begin{center}
    \includegraphics[width=0.35\textwidth]{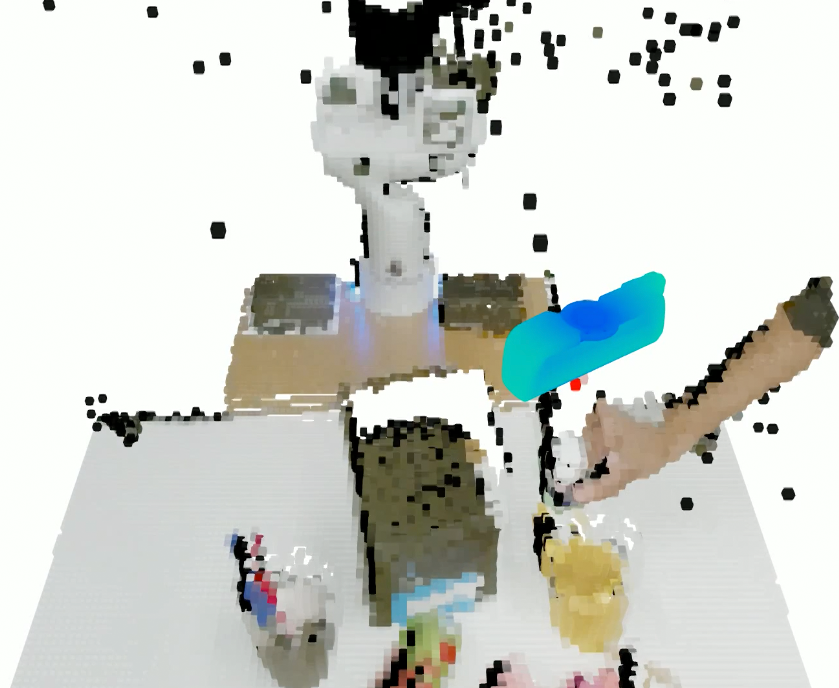}
  \caption{\textbf{Object Tracker.} Tracking  an unseen hand sanitizer instance.}
  \label{fig:tracking}
  \end{center}
\end{wrapfigure}
\subsection{Object Tracking}
Although \model~was not explicitly trained for 6-DoF object-tracking, our action detection framework can be used to localize objects in cluttered scenes. In this \href{https://peract.github.io/media/results/animations/handsan_tracking_v2.mp4}{video}, we show  an agent that was trained with one hand sanitizer instance on just 5 ``press the handsan'' demos, and then evaluated on tracking an unseen sanitizer instance. \model~does not need to build a complete representation of hand sanitizers, and only has to learn \textit{where to press} them. Our implementation runs at an inference speed of 2.23 FPS (or 0.45 seconds per frame), allowing for near real-time closed-loop behaviors. 

\begin{wrapfigure}{r}{0.40\textwidth}
  \vspace{-2.9cm}
  \begin{center}
    \includegraphics[width=0.31\textwidth]{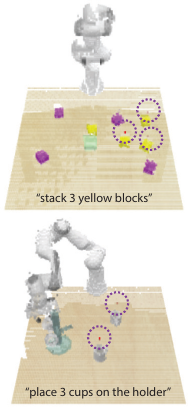}
  \caption{\textbf{Examples of Multi-Modal Predictions.}}
  \label{fig:multimodal}
  \end{center}
  \vspace{-1cm}
\end{wrapfigure}

\vspace{1cm}
\subsection{Multi-Modal Actions}
\model's problem formulation allows for modeling multi-modal action distributions, \ie scenarios where multiple actions are valid given a specific goal. \figref{fig:multimodal} presents some selected examples of multi-modal action predictions from \model. Since there are several \textit{``yellow blocks''} and \textit{``cups''} to choose from, the $\mathcal{Q}$-prediction distributions have several modes. In practice, we observe that the agent has a tendency to prefer certain object instances over others (like the front mug in \figref{fig:multimodal}) due to preference biases in the training dataset. We also note that the cross-entropy based training method from \secref{sec:training} is closely related to Energy-Based Models (EBMs)~\citep{lecun2006tutorial,florence2022implicit}. In a way, the cross-entropy loss is \textit{pulling up} expert 6-DoF actions, while \textit{pushing-down} every other action in the discretized action space. At test time, we simply maximize the learned $\mathcal{Q}$-predictions, instead of minimizing an energy function with optimization. Future works could look into EBM~\citep{florence2022implicit} training and inference methods for better generalization and execution performance.

\end{document}

%% file: tables/multi_task_results.tex
\begin{table}[!t]
\centering
\scriptsize

\vspace{-1.0cm}
\hspace*{-1.02cm}
\begin{tabular}{lcccccccccccccccccc} 
\toprule
                  & \multicolumn{2}{c}{\begin{tabular}[c]{@{}c@{}}\texttt{open} \\\texttt{drawer}\end{tabular}}     & \multicolumn{2}{c}{\begin{tabular}[c]{@{}c@{}}\texttt{slide} \\\texttt{block}\end{tabular}} &  \multicolumn{2}{c}{\begin{tabular}[c]{@{}c@{}}\texttt{sweep to} \\\texttt{dustpan}\end{tabular}}   &
                  \multicolumn{2}{c}{\begin{tabular}[c]{@{}c@{}}\texttt{meat off}\\\texttt{grill}\end{tabular}} &
                  \multicolumn{2}{c}{\begin{tabular}[c]{@{}c@{}}\texttt{turn} \\\texttt{tap}\end{tabular}}   & \multicolumn{2}{c}{\begin{tabular}[c]{@{}c@{}}\texttt{put in} \\\texttt{drawer}\end{tabular}}         & \multicolumn{2}{c}{\begin{tabular}[c]{@{}c@{}}\texttt{close}\\\texttt{jar}\end{tabular}}    &
                  \multicolumn{2}{c}{\begin{tabular}[c]{@{}c@{}}\texttt{drag} \\\texttt{stick}\end{tabular}}    &
                  \multicolumn{2}{c}{\begin{tabular}[c]{@{}c@{}}\texttt{stack}\\\texttt{blocks}\end{tabular}}  \\
                  \cmidrule(lr){2-3} \cmidrule(lr){4-5} \cmidrule(lr){6-7} \cmidrule(lr){8-9} \cmidrule(lr){10-11} \cmidrule(lr){12-13} 
                  \cmidrule(lr){14-15} \cmidrule(lr){16-17}
                  \cmidrule(lr){18-19}
                  \\[-13pt]                                                         \\
\vcell{Method}    & \vcell{10}       & \vcell{100}                                                & \vcell{10}       & \vcell{100}                                                      & \vcell{10}       & \vcell{100}                                                  & \vcell{10}       & \vcell{100}                                                 & \vcell{10}       & \vcell{100}                                           & \vcell{10}       & \vcell{100}                                                         & \vcell{10}       & \vcell{100}                                            & \vcell{10}       & \vcell{100}                                                  & \vcell{10}       & \vcell{100}                                             \\[-\rowheight]
\printcellbottom  & \printcellbottom & \printcellbottom                                           & \printcellbottom & \printcellbottom                                                 & \printcellbottom & \printcellbottom                                             & \printcellbottom & \printcellbottom                                            & \printcellbottom & \printcellbottom                                      & \printcellbottom & \printcellbottom                                                    & \printcellbottom & \printcellbottom                                       & \printcellbottom & \printcellbottom                                             & \printcellbottom & \printcellbottom                                        \\[0pt] 
\hline \\[-6pt]
\bczcnn         & 4                    & 4                                                      & 4                    & 0                                                            & 0                    & 0                                                        & 0                    & 0                                                          & 20                   & 8                                                & 0                    & 8                                                        & 0                    & 0                                                       & 0                    & 0                                                     & 0                    & 0                                                   \\
\highlight{\bczvit}   & 16                   & 0                                                  & 8                    & 0                                                  & 8                    & 0                                                       & 0                    & 0                                                     & 24                   & 16                                               & 0                    & 0                                                    & 0                    & 0                                                & 0                    & 0                                                 & 0                    & 0                                                   \\
\unet            & 28                   & 20                                                     & 12                   & 16                                                           & 4                    & 0                                                        & 40                   & 20                                                         & 60                   & 68                                               & 12                   & 4                                                        & 28          & 24                                                      & \textbf{72}          & 24                                                    & 4                    & 0                                                   \\
\model~(w/o Lang) & 20                   & 28                                                     & 8                    & 12                                                           & 20                   & 16                                                       & 40                   & 48                                                         & 36                   & 60                                               & 16                   & 16                                                       & 16                   & 12                                                      & 48                   & 60                                                    & 0                    & 0                                                   \\
\rowcolor[rgb]{0.9,1.0,0.9}\model           & \textbf{68}          & \textbf{80}                                            & \textbf{32}          & \textbf{72}                                                  & \textbf{72}          & \textbf{56}                                              & \textbf{68}          & \textbf{84}                                                & \textbf{72}          & \textbf{80}                                      & \textbf{16}          & \textbf{68}                                              & \textbf{32}          & \textbf{60}                                             & 36                   & \textbf{68}                                           & \textbf{12}          & \textbf{36}                                         \\[1pt]  
\hline \\
                  & \multicolumn{2}{c}{\begin{tabular}[c]{@{}c@{}}\texttt{screw}\\\texttt{bulb}\end{tabular}} & \multicolumn{2}{c}{\begin{tabular}[c]{@{}c@{}}\texttt{put in}\\\texttt{safe}\end{tabular}}      & \multicolumn{2}{c}{\begin{tabular}[c]{@{}c@{}}\texttt{place}\\\texttt{wine}\end{tabular}} & 
                  \multicolumn{2}{c}{\begin{tabular}[c]{@{}c@{}}\texttt{put in}\\\texttt{cupboard}\end{tabular}} &
                  \multicolumn{2}{c}{\begin{tabular}[c]{@{}c@{}}\texttt{sort}\\\texttt{shape}\end{tabular}} & 
                  \multicolumn{2}{c}{\begin{tabular}[c]{@{}c@{}}\texttt{push}\\\texttt{buttons}\end{tabular}}& \multicolumn{2}{c}{\begin{tabular}[c]{@{}c@{}}\texttt{insert}\\\texttt{peg}\end{tabular}} &
                  \multicolumn{2}{c}{\begin{tabular}[c]{@{}c@{}}\texttt{stack}\\\texttt{cups}\end{tabular}}         & \multicolumn{2}{c}{\begin{tabular}[c]{@{}c@{}}\texttt{place}\\\texttt{cups}\end{tabular}}    \\
        
                  \cmidrule(lr){2-3} \cmidrule(lr){4-5} \cmidrule(lr){6-7} \cmidrule(lr){8-9} \cmidrule(lr){10-11} \cmidrule(lr){12-13} 
                  \cmidrule(lr){14-15} \cmidrule(lr){16-17}
                  \cmidrule(lr){18-19}
                  \\[-6pt]          
\vcell{}          & \vcell{10}       & \vcell{100}                                                & \vcell{10}       & \vcell{100}                                                      & \vcell{10}       & \vcell{100}                                                  & \vcell{10}       & \vcell{100}                                                 & \vcell{10}       & \vcell{100}                                           & \vcell{10}       & \vcell{100}                                                         & \vcell{10}       & \vcell{100}                                            & \vcell{10}       & \vcell{100}                                                  & \vcell{10}       & \vcell{100}                                             \\[-\rowheight]
\printcellbottom  & \printcellbottom & \printcellbottom                                           & \printcellbottom & \printcellbottom                                                 & \printcellbottom & \printcellbottom                                             & \printcellbottom & \printcellbottom                                            & \printcellbottom & \printcellbottom                                      & \printcellbottom & \printcellbottom                                                    & \printcellbottom & \printcellbottom                                       & \printcellbottom & \printcellbottom                                             & \printcellbottom & \printcellbottom                                        \\[1pt]
\hline \\[-6pt]
\bczcnn          & 0                    & 0                                                      & 0                    & 4                                                            & 0                    & 0                                                        & 0                    & 0                                                          & 0                    & 0                                                & 4                    & 0                                                        & 0                    & 0                                                       & 0                    & 0                                                     & 0                    & 0                                                   \\
\highlight{\bczvit}   & 0                    & 0                                                  & 0                    & 0                                                  & 4                    & 0                                                       & 4                    & 0                                                     & 0                    & 0                                                & 16                   & 0                                                    & 0                    & 0                                                & 0                    & 0                                                 & 0                    & 0                                                   \\
\unet             & 12                   & 8                                                      & 0                    & 12                                                           & \textbf{36}          & 8                                                        & \textbf{4}           & 0                                                          & 8           & 8                                                & \textbf{88}          & \textbf{72}                                              & 0                    & \textbf{4}                                                       & 0                    & 0                                                     & 0                    & 0                                                   \\
\model~(w/o Lang) & 0                    & \textbf{24}                                            & 8                    & 20                                                           & 8                    & \textbf{20}                                                       & 0                    & 0                                                          & 0                    & 0                                                & 60                   & 68                                                       & 4                    & 0                                                       & 0                    & 0                                                     & 0                    & 0                                                   \\
\rowcolor[rgb]{0.9,1.0,0.9}\model~ & \textbf{28}          & \textbf{24}                                            & \textbf{16}          & \textbf{44}                                                  & 20                   & 12                                                       & 0                    & \textbf{16}                                                & \textbf{16}          & \textbf{20}                                      & 56                   & 48                                                       & \textbf{4}           & 0                                                       & 0                    & 0                                                     & 0                    & 0                                                   \\[-1pt]
\bottomrule
\end{tabular}
\vspace{2pt}
\caption{\textbf{Multi-Task Test Results.} Success rates (mean \%) of various multi-task agents tasks trained with either 10 or 100 demonstrations per task and evaluated on 25 episodes per task. Each evaluation episode is scored either a 0 for failure or 100 for succces. \model~outperforms \unet\citep{c2farm}, \highlight{the most competitive baseline, with an average improvement of $1.33\times$ with 10 demos and $2.83\times$ with 100 demos.}}
\vspace{-0.8cm}
\label{table:rlbench}
\end{table}

%% file: tables/real_robot_results.tex
\begin{wraptable}{r}{0.3\textwidth}
  \vspace{-1.3em}
  \setlength\tabcolsep{2.3pt}
  \centering
  \scriptsize
\begin{tabular}{lccc} 
\toprule
Task          & \# Train~ & \# Test & Succ. \%  \\ 
\midrule
Press Handsan & 5         & 10      & 90        \\
Put Marker    & 8         & 10      & 70        \\
Place Food    & 8         & 10      & 60        \\
Put in Drawer & 8         & 10      & 40        \\
Hit Ball      & 8         & 10      & 60        \\
Stack Blocks  & 10        & 10      & 40        \\
Sweep Beans   & 8         & 5       & 20        \\
\bottomrule
\end{tabular}
    \caption{\scriptsize{Success rates (mean \%) of a multi-task model trained an evaluated 7 real-world tasks (see \figref{fig:tasks}).}} %
  \vspace{-2em}
  \label{table:real}
\end{wraptable}

%% file: tables/task_desc.tex
\begin{table}[h]
\centering
\scriptsize
\begin{tabular}{llccl} 
\toprule
Task                      & Variation   Type           & \# of Variations     & Avg. Keyframes       & Language     Template         \\
\midrule
\texttt{open drawer}      & placement                  &           3          &           3.0        & ``open the \blank drawer'' \\
\texttt{slide block}      & color                      &           4          &           4.7        & ``slide the block to \blank target'' \\
\texttt{sweep to dustpan} & size                       &           2          &           4.6        & ``sweep dirt to the \blank dustpan'' \\
\texttt{meat off grill}   & category                   &           2          &           5.0        & ``take the \blank off the grill'' \\
\texttt{turn tap}         & placement                  &           2          &           2.0        & ``turn \blank tap'' \\
\texttt{put in drawer}    & placement                  &           3          &          12.0        & ``put the item in the \blank drawer'' \\
\texttt{close jar}        & color                      &          20          &           6.0        & ``close the \blank jar'' \\
\texttt{drag stick}       & color                      &          20          &           6.0        & ``use the stick to drag the cube onto the \blank target'' \\
\texttt{stack blocks}     & color, count               &          60          &          14.6        & ``stack \blank \blank blocks''  \\
\texttt{screw bulb}       & color                      &          20          &           7.0        & ``screw in the \blank light bulb'' \\
\texttt{put in safe}      & placement                  &           3          &           5.0        & ``put the money away in the safe on the \blank shelf'' \\
\texttt{place wine}       & placement                  &           3          &           5.0        & ``stack the wine bottle to the \blank of the rack'' \\
\texttt{put in cupboard}  & category                   &           9          &           5.0        & ``put the \blank in the cupboard'' \\
\texttt{sort shape}       & shape                      &           5          &           5.0        & ``put the \blank in the shape sorter'' \\
\texttt{push buttons}     & color                      &          50          &           3.8        & ``push the \blank button, [then the \blank button]'' \\
\texttt{insert peg}       & color                      &          20          &           5.0        & ``put the ring on the \blank spoke'' \\
\texttt{stack cups}       & color                      &          20          &          10.0        & ``stack the other cups on top of the \blank cup'' \\
\texttt{place cups}       & count                      &           3          &          11.5        & ``place \blank cups on the cup holder'' \\
\bottomrule
\end{tabular}
\vspace{0.2cm}
\caption{\scriptsize \textbf{Language-Conditioned Tasks in RLBench}~\citep{james2020rlbench}.}
\label{table:task_desc}
\end{table}

%% file: tables/sen_analysis.tex
\begin{table}[!b]
\centering
\scriptsize
\hspace*{-1cm}
\caption{\textbf{Sensitivity Analysis.} Success rates (mean \%) of various \model~agents trained with 100 demonstrations per task. We \\ investigate three factors that affect \model's performance: rotation  augmentation, number of Perceiver latents, and voxel resolution. }
\vspace{0.1cm}
\hspace*{-0.1cm}
\label{table:sen_analysis}
\begin{tabular}{lccccccccc} 
\toprule
                & \multicolumn{1}{c}{\begin{tabular}[c]{@{}c@{}}\texttt{open} \\\texttt{drawer}\end{tabular}} & \multicolumn{1}{c}{\begin{tabular}[c]{@{}c@{}}\texttt{slide} \\\texttt{block}\end{tabular}} & \multicolumn{1}{c}{\begin{tabular}[c]{@{}c@{}}\texttt{sweep to} \\\texttt{dustpan}\end{tabular}} & \multicolumn{1}{c}{\begin{tabular}[c]{@{}c@{}}\texttt{meat off}\\\texttt{grill}\end{tabular}}  & \multicolumn{1}{c}{\begin{tabular}[c]{@{}c@{}}\texttt{turn} \\\texttt{tap}\end{tabular}}  & \multicolumn{1}{c}{\begin{tabular}[c]{@{}c@{}}\texttt{put in} \\\texttt{drawer}\end{tabular}} & \multicolumn{1}{c}{\begin{tabular}[c]{@{}c@{}}\texttt{close}\\\texttt{jar}\end{tabular}}  & \multicolumn{1}{c}{\begin{tabular}[c]{@{}c@{}}\texttt{drag} \\\texttt{stick}\end{tabular}} & \multicolumn{1}{c}{\begin{tabular}[c]{@{}c@{}}\texttt{stack}\\\texttt{blocks}\end{tabular}}  \\ 
\cline{1-10} \\[-4pt]
\model~                & 80                                                    & 72                                                    & 56                                                         & 84                                                       & 80                                                  & 68                                                      & 60                                                  & 68                                                   & 36                                                     \\& \multicolumn{1}{l}{}                                  & \multicolumn{1}{l}{}                                  & \multicolumn{1}{l}{}                                       & \multicolumn{1}{l}{}                                     & \multicolumn{1}{l}{}                                & \multicolumn{1}{l}{} \\[-0.6em]
\rowcolor[rgb]{0.898,1.0,0.953} \model~ w/o Rot Aug  & 92                                                    & 72                                                    & 56                                                         & 92                                                       & 96                                                  & 60                                                      & 56                                                  & 100                                                  & 8                                                      \\
& \multicolumn{1}{l}{}                                  & \multicolumn{1}{l}{}                                  & \multicolumn{1}{l}{}                                       & \multicolumn{1}{l}{}                                     & \multicolumn{1}{l}{}                                & \multicolumn{1}{l}{} \\[-0.6em]
\rowcolor[rgb]{0.9,1.0,0.9} \model~ $4096$ latents & 84                                                    & 88                                                    & 44                                                         & 68                                                       & 84                                                  & 48                                                      & 48                                                  & 84                                                   & 12                                                     \\
\rowcolor[rgb]{0.9,1.0,0.9} \model~ $1024$ latents & 84                                                    & 48                                                    & 52                                                         & 84                                                       & 84                                                  & 52                                                      & 32                                                  & 92                                                   & 12                                                     \\
\rowcolor[rgb]{0.9,1.0,0.9} \model~ $\phantom{0}512$ latents  & 92                                                    & 84                                                    & 48                                                         & 100                                                      & 92                                                  & 32                                                      & 32                                                  & 100                                                  & 20                                                     \\
& \multicolumn{1}{l}{}                                  & \multicolumn{1}{l}{}                                  & \multicolumn{1}{l}{}                                       & \multicolumn{1}{l}{}                                     & \multicolumn{1}{l}{}                                & \multicolumn{1}{l}{} \\[-0.6em]
\rowcolor[rgb]{0.96,0.96,1} \model~ $\phantom{0}64^3$ voxels    & 88                                                    & 72                                                    & 80                                                         & 60                                                       & 84                                                  & 36                                                      & 40                                                  & 84                                                   & 32                                                     \\
\rowcolor[rgb]{0.96,0.96,1} \model~ $\phantom{0}32^3$ voxels    & 28                                                    & 44                                                    & 100                                                        & 60                                                       & 72                                                  & 24                                                      & 0                                                   & 24                                                   & 0                                                      \\& \multicolumn{1}{l}{}                                  & \multicolumn{1}{l}{}                                  & \multicolumn{1}{l}{}                                       & \multicolumn{1}{l}{}                                     & \multicolumn{1}{l}{}                                & \multicolumn{1}{l}{} \\[-0.6em]
\rowcolor[rgb]{1.0,1.0,0.90} \model~ $\phantom{0.}7^3$ patches    & 72                                                    & 48                                                    & 96                                                         & 92                                                       & 76                                                  & 76                                                      & 36                                                  & 96                                                   & 32                                                     \\
\rowcolor[rgb]{1.0,1.0,0.90} \model~ $\phantom{0.}9^3$ patches   &  68                                                    & 64                                                    & 56                                                         & 52                                                       & 96                                                  & 56                                                      & 36                                                  & 92                                                   & 20                                                     \\[2pt] 
\hline
                      &                                                                           &                                                                           &                                                                                &                                                                              &                                                                         &                                                                             &                                                                         &                                                                          &                                                                            \\
                      & \multicolumn{1}{c}{\begin{tabular}[c]{@{}c@{}}\texttt{screw}\\\texttt{bulb}\end{tabular}}   & \multicolumn{1}{c}{\begin{tabular}[c]{@{}c@{}}\texttt{put in}\\\texttt{safe}\end{tabular}}  & \multicolumn{1}{c}{\begin{tabular}[c]{@{}c@{}}\texttt{place}\\\texttt{wine}\end{tabular}}        & \multicolumn{1}{c}{\begin{tabular}[c]{@{}c@{}}\texttt{put in}\\\texttt{cupboard}\end{tabular}} & \multicolumn{1}{c}{\begin{tabular}[c]{@{}c@{}}\texttt{sort}\\\texttt{shape}\end{tabular}} & \multicolumn{1}{c}{\begin{tabular}[c]{@{}c@{}}\texttt{push}\\\texttt{buttons}\end{tabular}}   & \multicolumn{1}{c}{\begin{tabular}[c]{@{}c@{}}\texttt{insert}\\\texttt{peg}\end{tabular}} & \multicolumn{1}{c}{\begin{tabular}[c]{@{}c@{}}\texttt{stack}\\\texttt{cups}\end{tabular}}  & \multicolumn{1}{c}{\begin{tabular}[c]{@{}c@{}}\texttt{place}\\\texttt{cups}\end{tabular}}    \\[4pt]
\cline{1-10} \\[-4pt]
\model~                & 24                                                    & 44                                                    & 12                                                         & 16                                                       & 20                                                  & 48                                                      & 0                                                   & 0                                                    & 0                                                      \\& \multicolumn{1}{l}{}                                  & \multicolumn{1}{l}{}                                  & \multicolumn{1}{l}{}                                       & \multicolumn{1}{l}{}                                     & \multicolumn{1}{l}{}                                & \multicolumn{1}{l}{} \\[-0.6em]
\rowcolor[rgb]{0.898,1.0,0.953}\model~ w/o Rot Aug  & 20                                                    & 32                                                    & 48                                                         & 8                                                        & 8                                                   & 56                                                      & 8                                                   & 4                                                    & 0                                                      \\& \multicolumn{1}{l}{}                                  & \multicolumn{1}{l}{}                                  & \multicolumn{1}{l}{}                                       & \multicolumn{1}{l}{}                                     & \multicolumn{1}{l}{}                                & \multicolumn{1}{l}{} \\[-0.6em]
\rowcolor[rgb]{0.9,1.0,0.9}\model~ $4096$ latents & 32                                                    & 44                                                    & 52                                                         & 8                                                        & 12                                                  & 72                                                      & 4                                                   & 4                                                    & 0                                                      \\
\rowcolor[rgb]{0.9,1.0,0.9}\model~ $1024$ latents & 24                                                    & 32                                                    & 36                                                         & 8                                                        & 20                                                  & 40                                                      & 8                                                   & 4                                                    & 0                                                      \\
\rowcolor[rgb]{0.9,1.0,0.9}\model~ $\phantom{0}512$ latents  & 48                                                    & 40                                                    & 36                                                         & 24                                                       & 16                                                  & 32                                                      & 12                                                  & 0                                                    & 4                                                      \\& \multicolumn{1}{l}{}                                  & \multicolumn{1}{l}{}                                  & \multicolumn{1}{l}{}                                       & \multicolumn{1}{l}{}                                     & \multicolumn{1}{l}{}                                & \multicolumn{1}{l}{} \\[-0.6em]
\rowcolor[rgb]{0.96,0.96,1} \model~ $\phantom{0}64^3$ voxels    & 24                                                    & 48                                                    & 44                                                         & 12                                                       & 4                                                   & 32                                                      & 0                                                   & 4                                                    & 0                                                      \\
\rowcolor[rgb]{0.96,0.96,1} \model~ $\phantom{0}32^3$ voxels    & 12                                                    & 20                                                    & 52                                                         & 0                                                        & 0                                                   & 60                                                      & 0                                                   & 0                                                    & 0                                                      \\& \multicolumn{1}{l}{}                                  & \multicolumn{1}{l}{}                                  & \multicolumn{1}{l}{}                                       & \multicolumn{1}{l}{}                                     & \multicolumn{1}{l}{}                                & \multicolumn{1}{l}{} \\[-0.6em]
\rowcolor[rgb]{1.0,1.0,0.90} \model~ $\phantom{0.}7^3$ patches    & 8                                                     & 48                                                    & 76                                                         & 0                                                        & 12                                                  & 16                                                      & 0                                                   & 0                                                    & 0                                                      \\
\rowcolor[rgb]{1.0,1.0,0.90} \model~ $\phantom{0.}9^3$ patches    & 12                                                    & 36                                                    & 72                                                         & 12                                                       & 0                                                   & 20                                                      & 0                                                   & 0                                                    & 0                                                      \\
\bottomrule
\end{tabular}
\end{table}

%% file: tables/high_pres_res.tex
\begin{wraptable}{r}{0.3\textwidth}
\vspace{-0.89cm}
\centering
\scriptsize
\begin{tabular}{lcc} 
\toprule
              & Multi & Single  \\ 
\midrule
\texttt{place cups}    & 0     & 24     \\
\texttt{stack cups}    & 0     & 32      \\
\texttt{insert peg} & 0     & 16      \\
\bottomrule
\end{tabular}
\label{table:high_pres_res}
\caption{\scriptsize{Success rates (mean \%) of multi-task and single-task \model~agents trained with 100 demos and evaluated on 25 episodes.}}
\end{wraptable}